\documentclass[5p]{elsarticle}

\usepackage{textcomp, gensymb}
\usepackage[english]{babel}  
\usepackage{ragged2e, microtype}

\usepackage{amsmath} 
\usepackage{amssymb}  
\usepackage{verbatim}  
\usepackage{mathtools}
\usepackage{multirow}
\usepackage{booktabs}
\usepackage{graphicx}
 
\usepackage{xcolor,colortbl}
\usepackage{makecell} 
\usepackage{mwe} 
\usepackage{hhline}

\usepackage[caption=false,font=footnotesize]{subfig}

\definecolor{Gray}{gray}{0.85}
\definecolor{LightCyan}{rgb}{0.88,1,1}

\newcolumntype{a}{>{\columncolor{Gray}}c}

\begin{document}

\title{A Reactive performance-based Shared Control Framework for Assistive Robotic Manipulators}

\author[1,3]{Francisco J. Ruiz-Ruiz\corref{cor1}}
\ead{fjruiz2@uma.es}

\author[2]{Cristina Urdiales}
\ead{acurdiales@uma.es}

\author[2]{Manuel Fernández-Carmona}
\ead{mfcarmona@uma.es}

\author[1]{Jesús M. Gómez-de-Gabriel}
\ead{jesus.gomez@uma.es}

\affiliation[1]{organization={Robotics and Mechatronics Group, University of Málaga},
addressline={C/ Doctor Oriz Ramos, s/n},
postcode={29071},
city={Málaga},
country={Spain}}
\affiliation[2]{organization={Ingeniería de Sistemas Integrados Group, University of Málaga},
addressline={Bulevar Louis Pasteur, 35},
postcode={29071},
city={Málaga},
country={Spain}}
\affiliation[3]{organization={Human Robot Interfaces and Interaction Lab, Istituto Italiano di Tecnologia},
addressline={Via San Quirico, 19D},
postcode={16163},
city={Genova},
country={Italy}}

\cortext[cor1]{Corresponding author}

\begin{abstract}
In Physical Human--Robot Interaction (pHRI) grippers, humans and robots may contribute simultaneously to actions, so it is necessary to determine how to combine their commands. Control may be swapped from one to the other within certain limits, or input commands may be combined according to some criteria. The Assist-As-Needed (AAN) paradigm focuses on this second approach, as the controller is expected to provide the minimum required assistance to users. Some AAN systems rely on predicting human intention to adjust actions. However, if prediction is too hard, reactive AAN systems may weigh input commands into an emergent one. This paper proposes a novel AAN reactive control system for a robot gripper where input commands are weighted by their respective local performances. Thus, rather than minimizing tracking errors or differences to expected velocities, humans receive more help depending on their needs. The system has been tested using a gripper attached to a sensitive robot arm, which provides evaluation parameters. 
Tests consisted of completing an on-air planar path with both arms. After the robot gripped a person's forearm, the path and current position of the robot were displayed on a screen to provide feedback to the human. The proposed control has been compared to results without assistance and to impedance control for benchmarking. A statistical analysis of the results proves that global performance improved and tracking errors decreased for ten volunteers with the proposed controller. Besides, unlike impedance control, the proposed one does not significantly affect exerted forces, command variation, or disagreement, measured as the angular difference between human and output command. Results support that the proposed control scheme fits the AAN paradigm, although future work will require further tests for more complex environments and tasks.
\end{abstract}


\maketitle

\section{Introduction}
\label{sec:introduction}

Nowadays, there is an increasing interest in cooperation between humans and robots. Humans may provide situation awareness, logic, and problem-solving capability to robots, whereas robots may provide skills, precision, and strength that exceed those of a human~\cite{sicilianoSurvey}. Although cooperation does not necessarily imply physical contact between human and robot, assistive robots are expected to contribute to human motion and/or joint torque generation, e.g., rehabilitation or human augmentation~\cite{pons2008wearable}, and may require physical Human-Robot Interaction (pHRI). pHRI robots require extra effort in mechanical design and control methods due to safety constraints.

To achieve a common task, pHRI robots need to cooperate with humans, i.e., both robot and human contribute to emergent motion, relying on Shared Control (SC) approaches that may imply different levels of autonomy for involved agents. Original approaches to SC typically ranged from safeguard operation, where the robot takes over when hazardous situations are detected, to behavior selection, where control is swapped from human to machine to perform specific tasks. Sometimes, an arbitration system may choose whether humans or robots contribute to control at each time~\cite{losey2018review}. In pHRI SC, both agents have a (varying) impact on the emerging command at all times. Hence, an arbitration mechanism must adjust robot and human contribution to control in a continuous way~\cite{Ajoudani2018}. If the human contribution to control is neglected, humans may lose interest in the task, get upset and/or stressed, reject assistance, and, in extreme, be physically harmed. Furthermore, in assistive robots, excess assistance may lead to loss of residual skills. In this sense, minimizing robotic assistance in rehabilitating patients with partial control of their limbs has reportedly shown good results in inducing neural plasticity~\cite{wolbrecht2008optimizing}. A type of controller that exerts the minimum required assistance to the user, known as assist-as-needed (AAN), maximizes the benefit of user engagement~\cite{pehlivan}. 

In AAN systems, the robot must often dynamically predict the user's intent. In applications where roles are clearly defined, and targets are easy to identify, state-based models can be used~\cite{jeon:20}. On the other hand, some approaches rely on assigning a set of skills to users and adapting assistance to their detected skill level rather than on human intentions~\cite{enayati:18}. 
Finally, if human intention is too hard to predict, SC can reactively adjust depending on feedback errors~\cite{li:15}. 

This paper tackles the challenge of AAN for a robot arm with force-sensing capability. The proposed control framework is an evolution from previous research by the authors on SC for smart wheelchairs~\cite{sharedControlCurdiales}. However, this approach required a significant transformation because i) in wheelchairs, users provide direct commands via a control device, whereas in this case, intention must be extracted from interaction; and ii) emergent forces have a direct physical impact on users, whereas in wheelchairs assistance mostly affected emergent trajectories. 

In this work, the robot assists the user in following a (proposed) predefined path known to both the robot and the user. No force nor speed constraints are imposed except an upper safety bound. During physical contact, interaction force is used to determine the user's intention. This command is combined with an analytically calculated one toward the goal if the robot operated independently. Both commands are then weighted by their respective local performances and combined into a single emergent one. Hence, the robot corrects human commands more or less depending on their respective performance. As the device that measures human input and the robot end-effector (EE) are the same, inputs depend on robot motion, as in a force-feedback bilateral teleoperation system. 
The proposed SC system lets humans contribute as much as possible to motion according to their performance.

The main difference between the proposed SC method and previous AAN approaches is that most methods typically focus on reducing tracking errors~\cite{Cherubini2016}. In~\cite{pehlivan}, the authors propose a minimal AAN control based on a baseline PD controller that compensates user inputs, estimated through a model-disturbance observer, providing an adjustable tracking error of a reference trajectory. 
Impedance controllers have also been used for implementing AAN controllers~\cite{Mihelj,zhang_AAN} due to their inherent compliance. Impedance-based AAN controllers minimize tracking error by incorporating a torque-free area around/along the reference trajectory/point, and virtual elements that restrain user actions in other directions~\cite{Kana2021,Wang2022}. Such controllers implement also tangential forces to force the users to follow the trajectory. To increase timing freedom, SC methods in the velocity domain~\cite{cervantes} aim at defining a desired velocity at every point of the task space and compensating for user input to achieve the desired trajectory, though they often compel the robot to track the desired velocities instead~\cite{Luo2020}. In all works discussed above, the unknown dynamics of the system need to be modeled to cancel undesired input properly. For example, in~\cite{Li2017} integral reinforcement learning (IRL) has been proposed to overcome the lack of information of the human arm model.
In the proposed method, human and robot provide local commands independently to follow the target path given their current position. Humans implicitly consider their own motion constraints and preferences when providing their command. Similarly, the robot motion generator considers robot constraints and targets but does not rely on any human model to provide a command. Performance-based command combination is not meant to minimize tracking errors but to favor commands that better preserve the properties of a navigation function, i.e., smoothness, safety, and directness. Our approach is reactive, so no models are required, plus emerging trajectories tend to preserve  

The main contributions of this paper are the following: 
\begin{itemize}
    \item A reactive velocity-based SC framework to assist robot-grasped human arms in completing a path.
    \item The improvement of human performance on an AAN basis, homogenizing performance among people with different skill levels but preserving their individual traits.
    \item An experimental evaluation framework to quantitatively compare the proposed approach with~\cite{zhang_AAN}, the most related AAN of the current state-of-the-art. 
\end{itemize}

The paper is structured as follows: in Section~\ref{sec:framework_overview}, the proposed framework is presented and applied to the control of a sensitive manipulator using interaction forces and velocity references. Sec.~\ref{sec:experiments} describes the experiments conducted and reports the results, which are subsequently discussed in Sec.~\ref{sec:results}. Finally, conclusions are synthesized in Sec.~\ref{sec:Conclusion}.

\section{A Reactive Shared Control Framework}
\label{sec:framework_overview}

\subsection{Framework Overview}

Robot-initiated pHRI requires careful mechanical design of effectors. Although this work focuses on SC, the following elements are required: i) a velocity-controlled robotic manipulator with force sensing capabilities; and ii) a gripper capable of providing a firm but compliant grasp (in this work, the gripper presented by the authors in~\cite{RUIZRUIZ_MaMT} is used).
Safety is ensured in the different components of this robot system. On the one hand, modern collaborative robots provide built-in safety mechanisms such as collision detection or speed limiting. On the other hand, the properties of the required gripper provide extra safety, allowing humans to release their arms if the robot security system fails. 

Many recent SC approaches control equations rely on linear blend (LB) of the user’s input and robot’s computed direction~\cite{ezeh:17}. SC performance may be improved by probabilistic models of user behavior and/or blending policies. However, without user models, LB remains the most popular choice. Hence, the proposed SC signal ($h$) relies on a linear combination of human ($u_h$) and robot control ($u_a$) signals, weighted by an $\alpha$ factor:
\begin{equation}
    \begin{array}{lr}
        h(u_h,u_a)=\alpha u_h +(1 - \alpha)u_a, & 0 \leq \alpha \leq 1.
    \end{array}
    \label{eq:comblin}
\end{equation}

In the proposed control framework, $u_h$ is extracted from interaction forces in the gripper. Simultaneously, a path follower continuously computes $u_a$ to reach the next target goal from the current location. 
It must be observed that $u_a$ does not depend on $u_h$: the robot always proposes its own way of reaching the next goal. The weight of each agent in the emerging action is provided by factor $\alpha$, and it could be manually set depending on the expected amount of assistance required or automatically estimated, e.g., by human intent recognition~\cite{jain:19}. However, in our approach, $\alpha$ is computed in terms of how appropriate human and robot commands are to achieve the desired goal ($\alpha = \eta(t)$), as proposed by the authors in~\cite{sharedControlCurdiales} for assisted wheelchair navigation.

To steadily adapt to the specific needs of users at each time instant, (\ref{eq:comblin}) needs to be calculated continuously, and, hence, performance needs to be locally obtained. In~\cite{sharedControlCurdiales}, local performance calculation was based on the properties of a navigation function, e.g., smoothness, safety, and directness, meaning that commands that preserve safety and heading and lead the vehicle to its (local) goal yield higher performance. However, as commented, the approach in \cite{sharedControlCurdiales} can not be directly extended to this framework because: i) contact is more complex and requires further adaptation, e.g., some grips may be uncomfortable to humans, and slippage depends on motion direction; and ii) agents physically affect each other, i.e., emergent commands need to account for exerted human force. 

\begin{figure}[!t]
    \centering
    \includegraphics[width=\columnwidth]{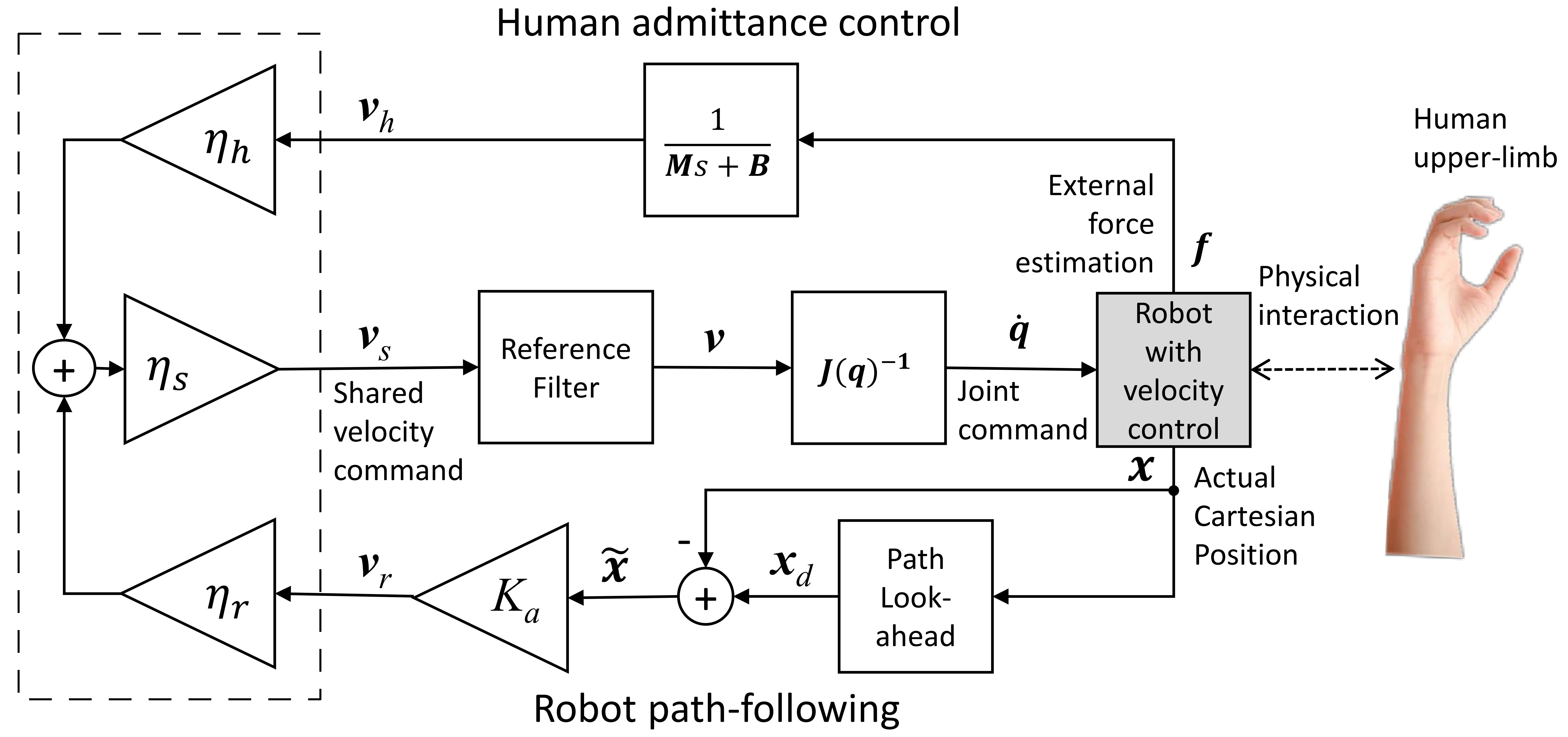}
    \caption{Block diagram of the proposed SC framework.}
    \label{fig:blocks}
\end{figure}

Thus, we propose the new SC in Figure~\ref{fig:blocks} to account for this increased physical feedback.
The proposed SC scheme works in a reactive, bottom-up way. In brief, given a sequence of partial goals composing a desired path at any given location, the robot calculates a speed command to reach the desired (partial) goal. Simultaneously, the human volunteer moves their arm to track the agreed upon path according to their preferences, and interaction forces are used by the system to predict their intention. The system computes the performance of both robot and human command vectors, weights them by their own performance, and combines them into a single emergent one, which is applied to the robot arm. Humans receive haptic feedback through this process, i.e., they can increase their force to move away from the partial goal, but they will perceive an increased resistance towards their actions.

\subsection{Computation of the Shared Velocity Command}

Let $\boldsymbol{x}\in \mathbb{R}^3$ be the EE position and let $\boldsymbol{x}_d \in \mathbb{R}^3$ be the goal point Cartesian coordinates. The proposed human command, $\boldsymbol{v}_h \in \mathbb{R}^3$, to reach $\boldsymbol{x}_d$ is extracted from the force applied at the EE. For the computation of the force-velocity conversion, an admittance-type control system is used, i.e. the dynamic behavior of a virtual passive mechanical system is rendered such as
\begin{equation}
    \boldsymbol{V}_h(s) = \frac{\boldsymbol{F}(s)}{\boldsymbol{M}s + \boldsymbol{B}}
    \label{eq:humaninput}
\end{equation}
where $\boldsymbol{M} \in \mathbb{R}^{3 \times 3}$ and $\boldsymbol{B} \in \mathbb{R}^{3 \times 3}$ are diagonal positive matrices denoting the virtual mass and damping of the system, $s$ is the Laplace variable, and $\boldsymbol{F}(s)$ and $\boldsymbol{V}(s)$ are the Laplace transforms of the human Cartesian force $\boldsymbol{f} \in \mathbb{R}^3$ and $\boldsymbol{v}_h$ respectively.

On the other hand, a proportional control computes the robot's proposed velocity command $\boldsymbol{v}_r \in \mathbb{R}^3$ as 
\begin{equation}
    \boldsymbol{v}_r = \boldsymbol{K}_{a} \tilde{\boldsymbol{x}},
    \label{eq:pfa}
\end{equation}
where $\boldsymbol{K}_{a} \in \mathbb{R}^{3 \times 3}$ is a diagonal positive matrix denoting the attraction constant to the goal, and $\tilde{\boldsymbol{x}} = \boldsymbol{x}_d - \boldsymbol{x}$ denotes the position error. $\boldsymbol{K}_{a}$ modulates the magnitude of the robot's action for a given position error. If the value of $\boldsymbol{K}_a$ is very high, the user will be dragged by the robot; if $\boldsymbol{K}_a$ is too low, the user will not receive assistance at all.  
Although more complex planning algorithms might be used, due to the reactive nature of this system, the system will not track the path fully predictively, so a proportional method fits our needs.

Once the robot and human have provided a command to reach the target position, the controller must combine them to obtain the Shared Velocity Command (SVC), i.e., the final command the robot will execute.
The SVC computation is performed in two phases. Firstly, the raw SVC, $\hat{\boldsymbol{v}}_s$, is obtained as follows:  
\begin{equation}
    \hat{\boldsymbol{v}}_s = \eta_r \boldsymbol{v}_r + \eta_h \boldsymbol{v}_h ,
    \label{eq:normsum}
\end{equation}

where $\eta_r$ and $\eta_h$ denote the performance of robot and human proposed commands. 

In the second phase, $\hat{\boldsymbol{v}}_s$ is weighted by its associated performance, $\eta_s$.
Thus, the final SVC, $\boldsymbol{v}_s$, is computed as
\begin{equation}
            \boldsymbol{v}_s = \eta_s \hat{\boldsymbol{v}}_s .
    \label{eq:modulo_as}
\end{equation}

\begin{figure*}[thp]
    \captionsetup[subfloat]{captionskip=1pt}
    \centering  
    \subfloat[\label{fig:pathfollower:a}]{%
    \includegraphics[width=0.33\textwidth]{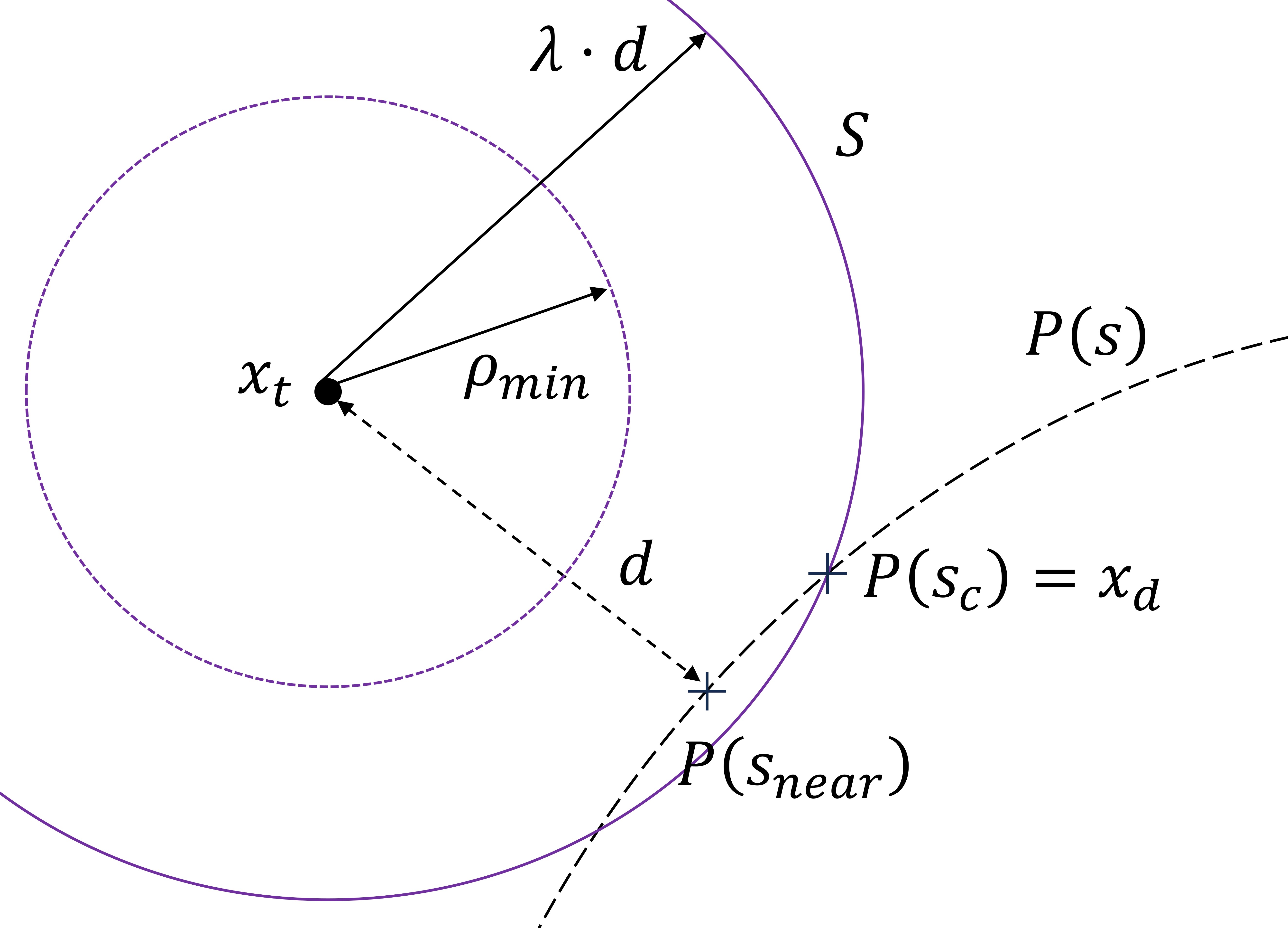}}
  \hfill
  \subfloat[\label{fig:pathfollower:b}]{%
    \includegraphics[width=0.33\textwidth]{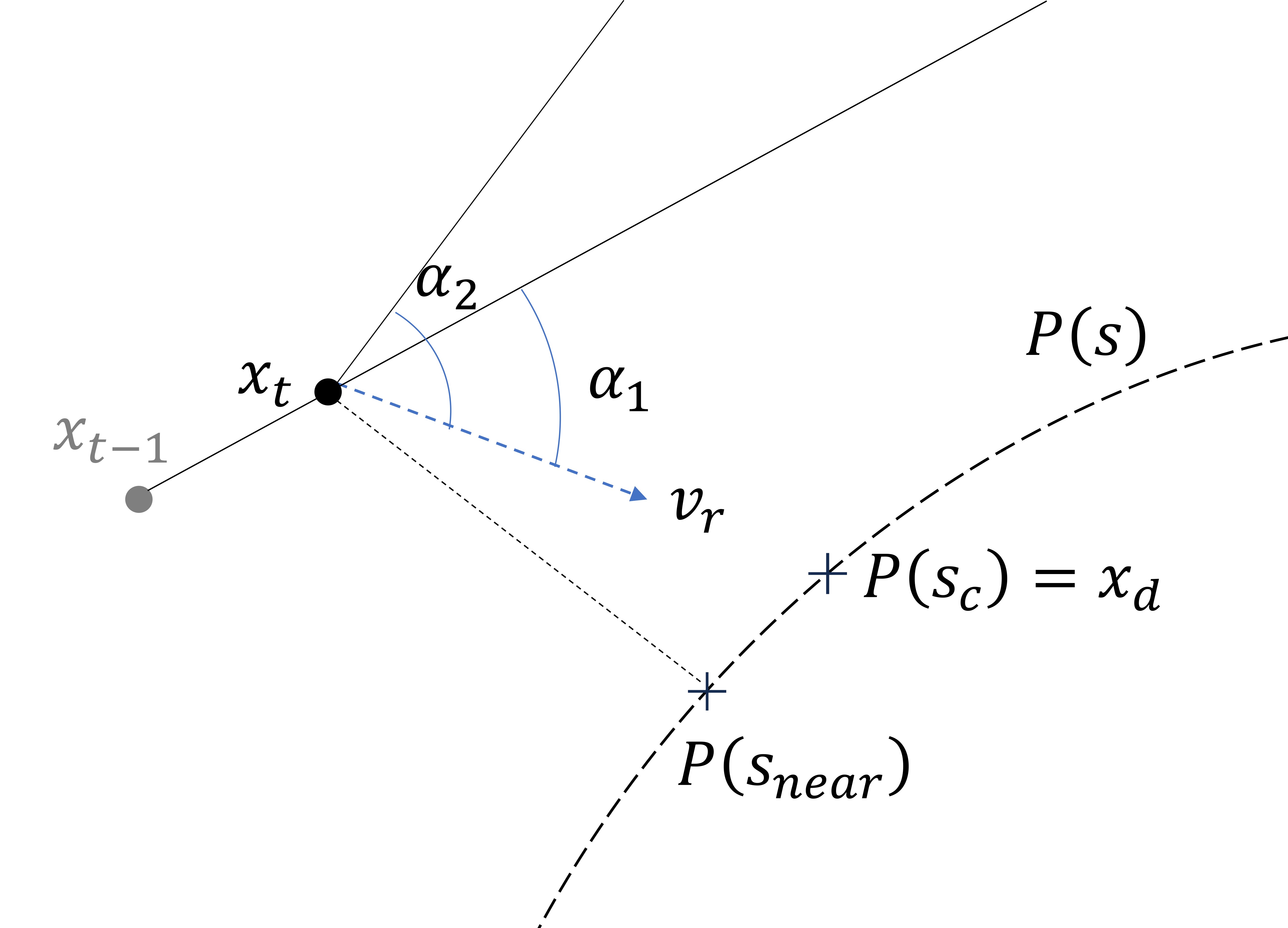}}
  \hfill
  \subfloat[\label{fig:pathfollower:c}]{%
    \includegraphics[width=0.33\textwidth]{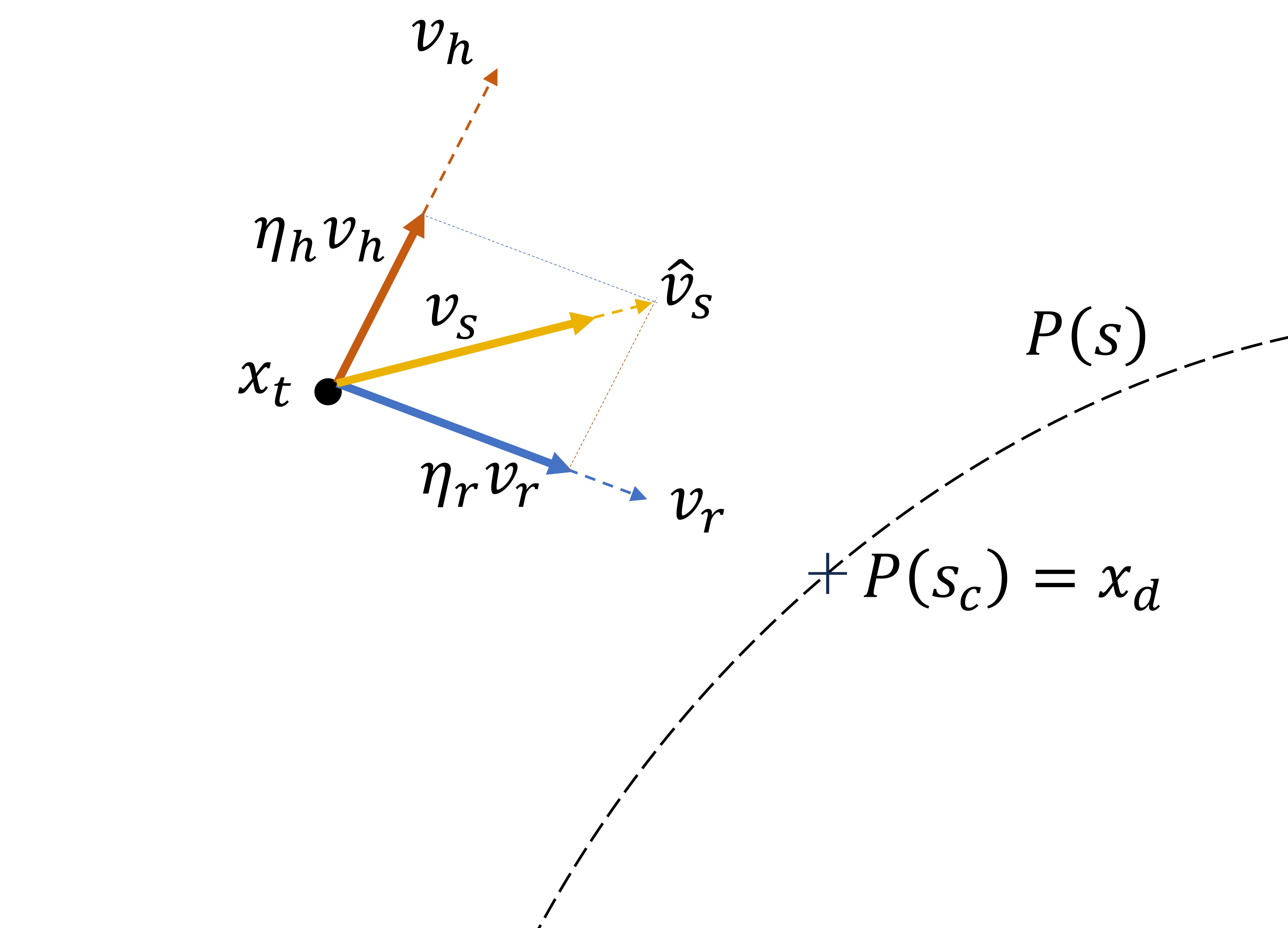}}
  \hfill
    \caption{Diagram 2D of the proposed reactive shared control scheme: a) The reactive path follower selects as the new objective point the one that intersects the virtual sphere $\mathcal{S}$ and makes the system advance in the positive direction of the path; b) Computation of the angles $\alpha_1$ and $\alpha_2$ for performance computation; c) combination of the proposed human and robot commands to generate the shared command.}
    \label{fig:pathfollower_scheme}
\end{figure*}

Figure~\ref{fig:pathfollower:c} illustrates the combination of human and robot commands to generate SVC. It must be noted that, while in the traditional convex blending law in (\ref{eq:comblin}) inputs are complementary, in (\ref{eq:normsum}) $\eta_r$ and $\eta_h$ are independent. As performance is task-oriented, if both commands present high performance, they are equally good at achieving the desired goal, and it usually means that they are both very similar, even if one may slightly favor smoothness and the other directness. However, if both commands present low performance, none is favored. As a result, emergent speed decreases. This braking effect is perceived as increased resistance against the input command by users, who tend to react by correcting their low efficient inputs. This pseudo-haptic effect helps to adapt to desired trajectories progressively.

Finally, to avoid sharp acceleration profiles, $\boldsymbol{v}_s$ is filtered by a low-pass filter and saturated if the norm of the SVC is bigger than the maximum Cartesian velocity allowed at the EE, $v_{max}$. 

\subsection{Performance Computation}
\label{subsec:eta}
The proposed SC works reactively, i.e., it relies on local performance assessment. Hence, it can not involve environment models or temporal information. Instead, it considers how well a given command fulfills a set of $m$ performance targets. Each performance target (e.g., smoothness, safety, precision, comfort...) is associated with a partial performance factor. Then, all these factors are averaged to obtain total performance. In brief, at any time instant, performance $\eta_k \in [0,1]$ associated with any given command $k$ (with $k=h,r,s$) can be obtained as the weighted combination of all $m$ partial performance factors $\eta_{k_i} \in [0,1]$ (with $i = 1, 2, ..., m$) considered:   
\begin{equation}
    \eta_k = \frac{\sum_{i=1}^m w_i \eta_{i}}{ \sum_{i=1}^m w_i} ,
    \label{eq:etatotal}
\end{equation}
where $w_i$ denotes the weight of performance factor $i$. By default, all weights are assumed to be equal so that all performance factors have the same importance. However, if a given task has specific requirements, like extra smoothness or precision, $w_i$ allows to favor one performance factor over others.

The task in this work is following a path,  so local performance calculation can be based on the main properties of a navigation function, as discussed by the authors in~\cite{fernandez:15}. 
Desired performance targets include safety, smoothness, and directness. However, in the present work, there are no obstacles in the test environment, and safety is hard-coded in the robot, so only two properties are considered ($m=2$): smoothness and directness. The smoothness factor of a command $\eta_{k_1}$, penalizes abrupt changes in the movement direction, so commands providing a smooth path will have higher performance. The directness factor $\eta_{k_2}$ favors commands that bring the system closer to its target, favoring trajectories that keep goals ahead. The smoothness and directness performance factors of a general command can be obtained using (\ref{eq:smooth_factor}) and (\ref{eq:direct_factor}), respectively: 
\begin{align}   
    \eta_{k_1} = e^{-C_{1} | \alpha_{k_1} | }, \label{eq:smooth_factor}\\
    \eta_{k_2} = e^{-C_{2} | \alpha_{k_2} | }, \label{eq:direct_factor}
\end{align}
where $C_{1}$ and $C_{2}$ are constants that modulate the slope of the importance of each factor, $\alpha_{k_1}$ denotes the angle between the previous executed command and proposed command $k$, and $\alpha_{k_2}$ represents the angle between a unitary vector tangent to the path in the current position and command $k$, Figure~\ref{fig:pathfollower:b}. As in the case of $w_i$, $C_i$ can be fixed to 1 or set -heuristically or otherwise, to a specific value that fits the problem at hand.

\subsection{Path Following}

In traditional motion planning, the robot computes a path from one point to another. Instead, the proposed framework considers a predefined path to be followed, but due to the influence of the human in the robot's motion, the robot will follow a path slightly different from the predefined one to make the path following reactive, a new objective point is dynamically chosen each time step.

Let us consider a virtual sphere $\mathcal{S}$ of dynamic radius $\rho$ centered at $\boldsymbol{x}$, and let $\mathcal{P}(s)$ be a 3D Cartesian path to be followed parameterized by the variable $s \in [0, 1]$. The robot deems the points of $\mathcal{P}(s)$ contained in the boundary of $\mathcal{S}$ as objective points candidates. To choose an objective point that makes the robot advance in the positive direction of the path, $\rho$ is computed so that $\mathcal{S}$ intersects with $\mathcal{P}(s)$ in at least two points, 
\begin{equation}
    \rho = \left\{ \begin{array}{lr}
            \lambda d , & \textrm{if }d \geq \rho_{min} \\
            \rho_{min} , & \textrm{Otherwise}
            \end{array} \right. ,
\end{equation}
where $\lambda > 1$ is the sphere dilation index, $\rho_{min}$ is the minimum radius of $\mathcal{S}$, and $d$ is the minimum distance between the actual robot position and the path obtained as 
\begin{equation}
    d = \text{dist}(\boldsymbol{x}, \mathcal{P}(s_{near})), 
\end{equation}
where the operator $\text{dist}(\cdot)$ is the Euclidean distance between two points, and $s_{near}$ is given by
\begin{equation}
    s_{near} = \underset{s}{\text{argmin}} \ \text{dist}(\boldsymbol{x}, \mathcal{P}(s)).
\end{equation}
The purpose of $\lambda$ is to ensure that $\mathcal{S}$ intersects with $\mathcal{P}(s)$ at every possible EE position, whereas $\rho_{min}$ prevent $\rho$ to obtain values close to $0$. The intersection point between $\mathcal{P}(s)$ and $\mathcal{S}$ that makes the robot advance in the positive direction of $s$ can be obtained as follow:
\begin{equation}
    s_c = \underset{s > s_{near}}{\text{argmin}} \ |\text{dist}(\boldsymbol{x}, \mathcal{P}(s)) - \rho| .
\end{equation}

Then, the goal point can be defined as $\boldsymbol{x}_d = \mathcal{P}(s_c)$. Figure~\ref{fig:pathfollower:a} shows a 2D scheme of the functioning of the path follower. It can be seen the importance of properly selecting $\lambda$ because this parameter affects directly the way the path is followed. Please, note that only a lower bound for $\lambda$ is provided, the maximum value of $\lambda$ is path dependent. In practice, $\lambda$ can be adjusted heuristically, although it is recommended to use values close to 1 to avoid filtering important details of the path. On the other hand, $\rho_{min}$ becomes relevant when the tracking error is low. It can be understood as how far the follower sets the objective under low tracking error. The election of $\rho_{min}$ is hardly dependent of the task and path considered.

\subsection{Stability}

As shown in Figure \ref{fig:blocks}, the interaction with the user follows an admittance-control scheme, which is a type of control that allows a robot with non-reversible actuators to follow external forces. It measures (or estimates) external forces and computes the desired position (or velocities) using the dynamic response of a virtual object in response to external forces. A robot control loop then achieves the desired velocities. Conversely, impedance control uses the effects of the external forces (variation in position or velocities) on a robot with reversible actuators to compute the robot joint torques without force sensing. The latter is suitable to control the interaction between the robot and a stiff environment, and the former control performs better when the robot interacts with a soft environment.

In this case, the control system y comprises a path-following control with a proportional control law, which can be tuned to provide stable path tracking and an admittance human control loop. It is hard to probe the stability of human admittance control under any user interaction (The environment) because there are no universal models for human behavior. For instance, when a human interacts with an unstable robot, he increases his arm stiffness, trying to stabilize the system, which is not good for an admittance controller.

To solve this problem, some studies use analytic models of the operator's arm to evaluate the stability numerically, but it is difficult to accurately estimate the stiffness of the human arm (the environment)\cite{duchaine2008} beforehand (offline). Other approaches propose an online stability index \cite{dimeas2016}, based on the frequency of the signals, which are an indication of the environment stiffness, so the parameters of the admittance control (i.e., M and B) are modified based on this index to achieve stability by reducing the gain of the control loop.

In the proposed control scheme, the loop gains are modulated through different performance indexes, calculated inline, which provide a better indication of the performance of the task. This set of indexes $\eta_{i} \in [0,1]$ changes the loop gains for both robot path-following and human admittance loops. The gains are the performance indexes, which are reduced when they detect a loss of controller performance in goal-related measurements (e.g., directness, smoothness, \...). This method overcomes the poor performance of conservative thresholds of passivity-based approaches. Additionally, the desired velocity reference includes a low-pass filter (plus a saturation) that increases the stability, considering that a human arm's bandwidth is low (under 2Hz) \cite{perreault2001effects}.

\section{Experiments and Results}
\label{sec:experiments}

\subsection{Experimental Setup}\label{sec:setup}

In order to test the proposed SC approach, the underactuated gripper proposed by the authors in~\cite{RUIZRUIZ_MaMT} was attached to a Franka Research 3 robot arm (Franka Emika GmbH, Germany), capable of compliant motion control and 6-axis force sensing. A PC with a real-time OS compatible with ROS Noetic was used to control the robot at a frequency of 1 kHz. The Cartesian velocity controller provided by the manufacturer of the robot was used. During tests, the robot grips a volunteer arm to assist them in following a predefined path. Robot commands are analytically calculated to follow the path precisely. Safety is hardcoded in the robot: it stops in case of excessive force/velocity or potential collision.
Human commands are estimated from interaction forces in the gripper, which the Franka Research 3 manipulator directly provides. As no physical interaction with the environment is produced, it can be assumed that all the forces detected by the robot are human inputs to the system. Since robots typically outperformed humans in precision, volunteers were expected to improve their tracing performance collaboratively. Consequently, we chose a circular path for this work, where users are expected to keep a constant curvature the whole time and close to the predefined path.

\begin{figure}[!t]
\centering
\includegraphics[width=\columnwidth]{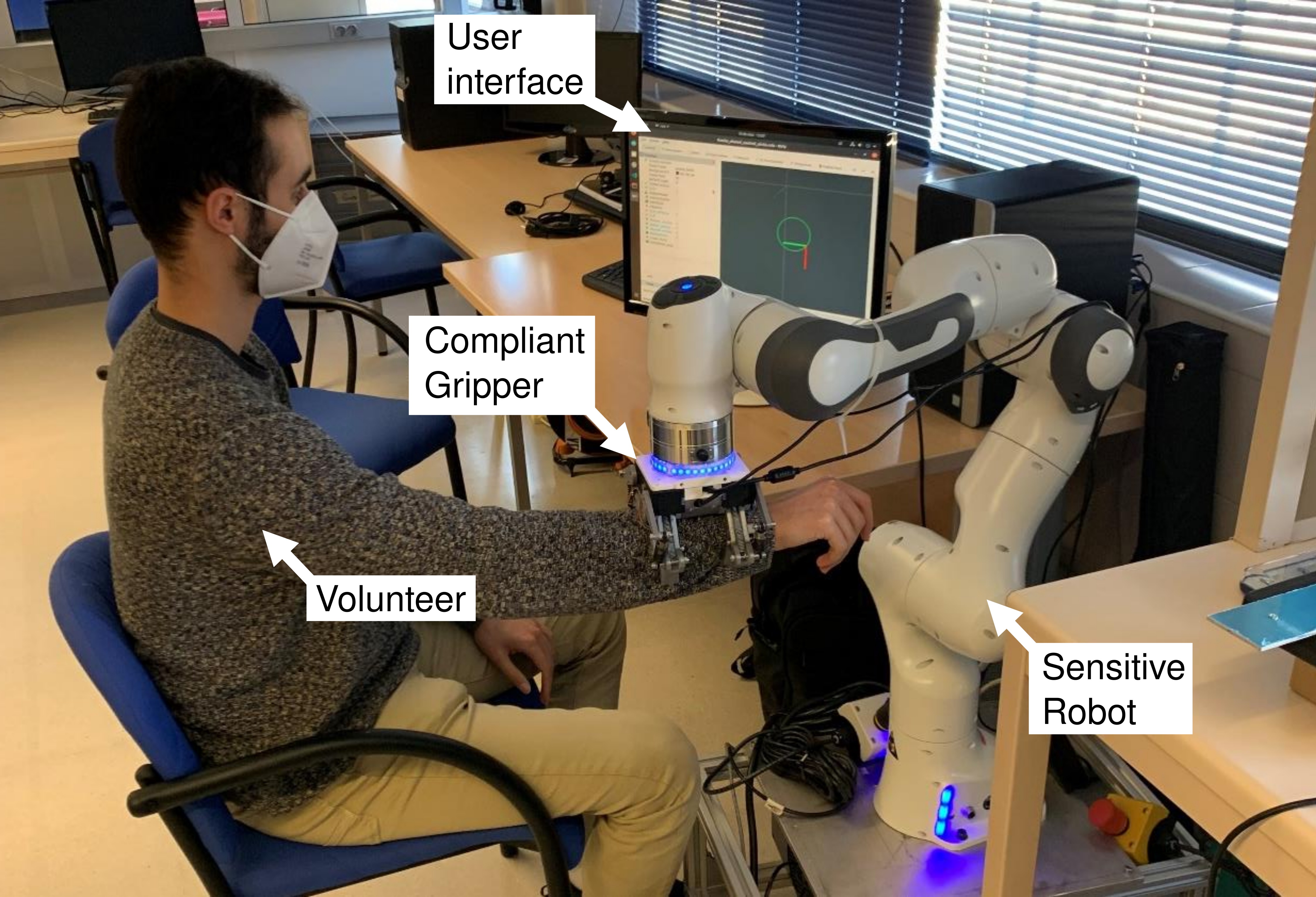}
\caption{Experimental setup. A volunteer performs a circular path with his right arm while grasped by a robot equipped with a compliant gripper. The volunteer receives real-time feedback about the path and their current position.}
\label{fig:setup}
\end{figure}

\begin{figure*} [!th]
  \captionsetup[subfloat]{captionskip=1pt}
    \centering  
    \subfloat[\label{fig:fran:traj:DA}]{%
    \includegraphics[width=0.33\textwidth]{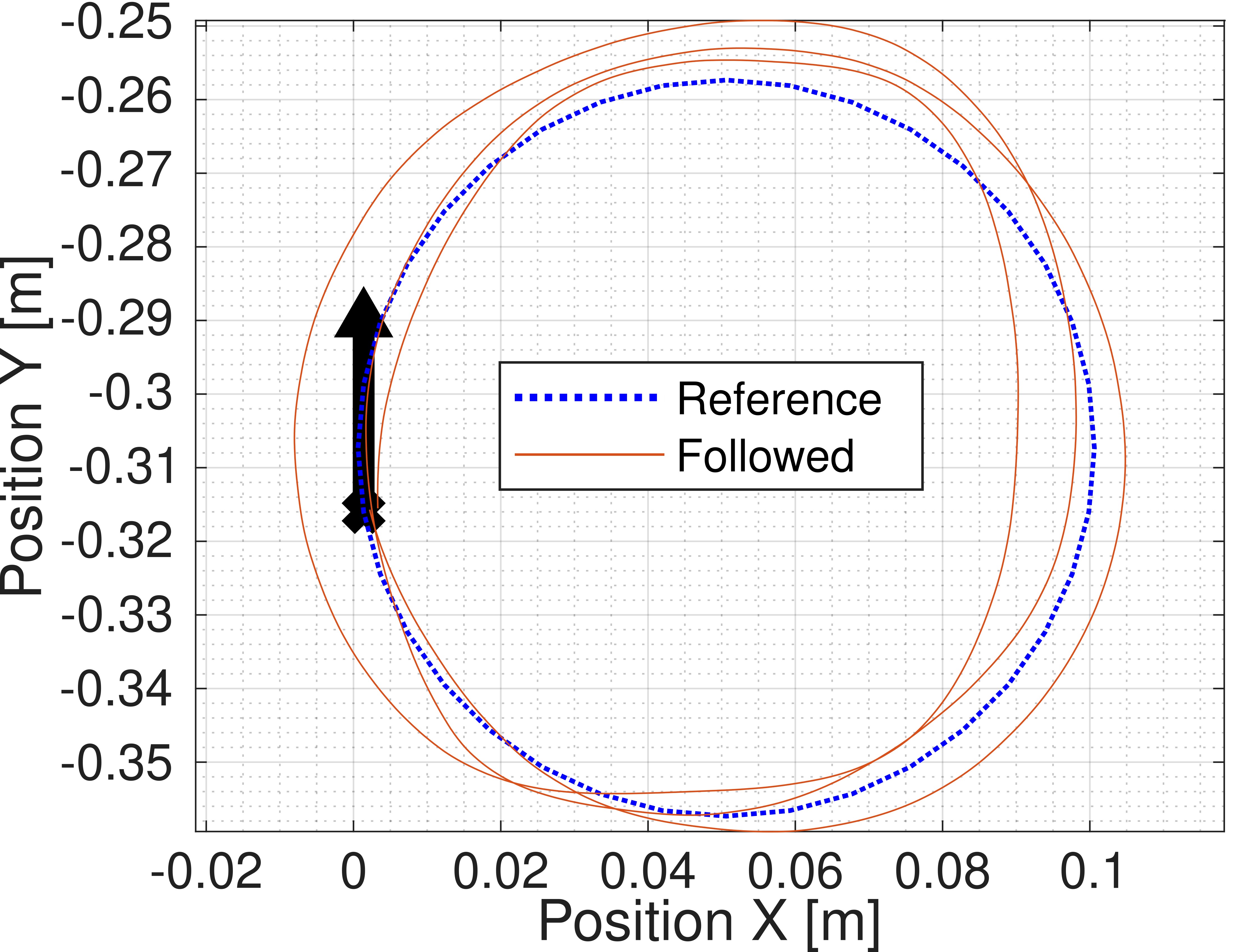}}
  \hfill
  \subfloat[\label{fig:fran:traj:DS}]{%
    \includegraphics[width=0.33\textwidth]{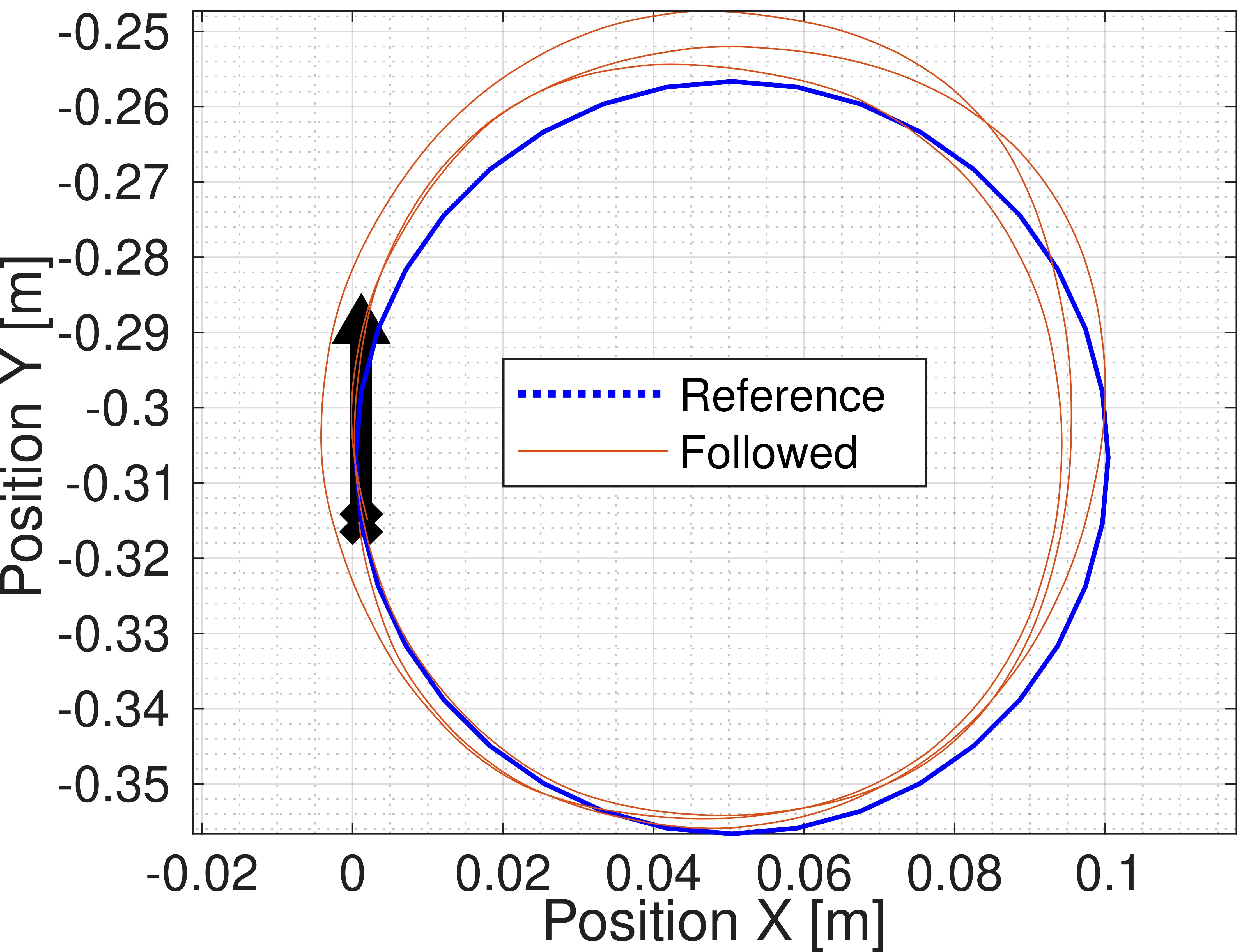}}
  \hfill
  \subfloat[\label{fig:fran:traj:DI}]{%
    \includegraphics[width=0.33\textwidth]{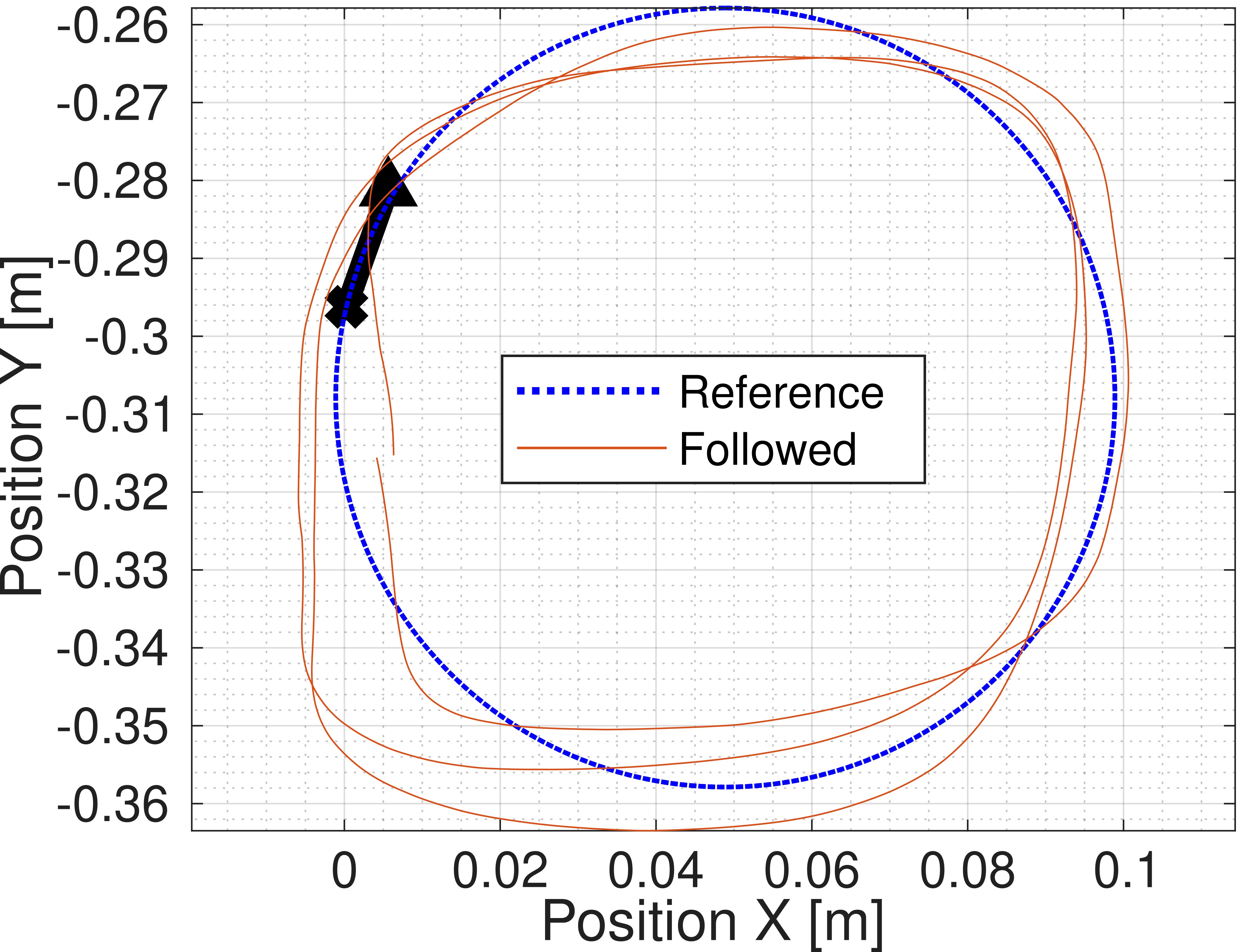}}
  \hfill
  \subfloat[\label{fig:fran:traj:NA}]{%
    \includegraphics[width=0.33\textwidth]{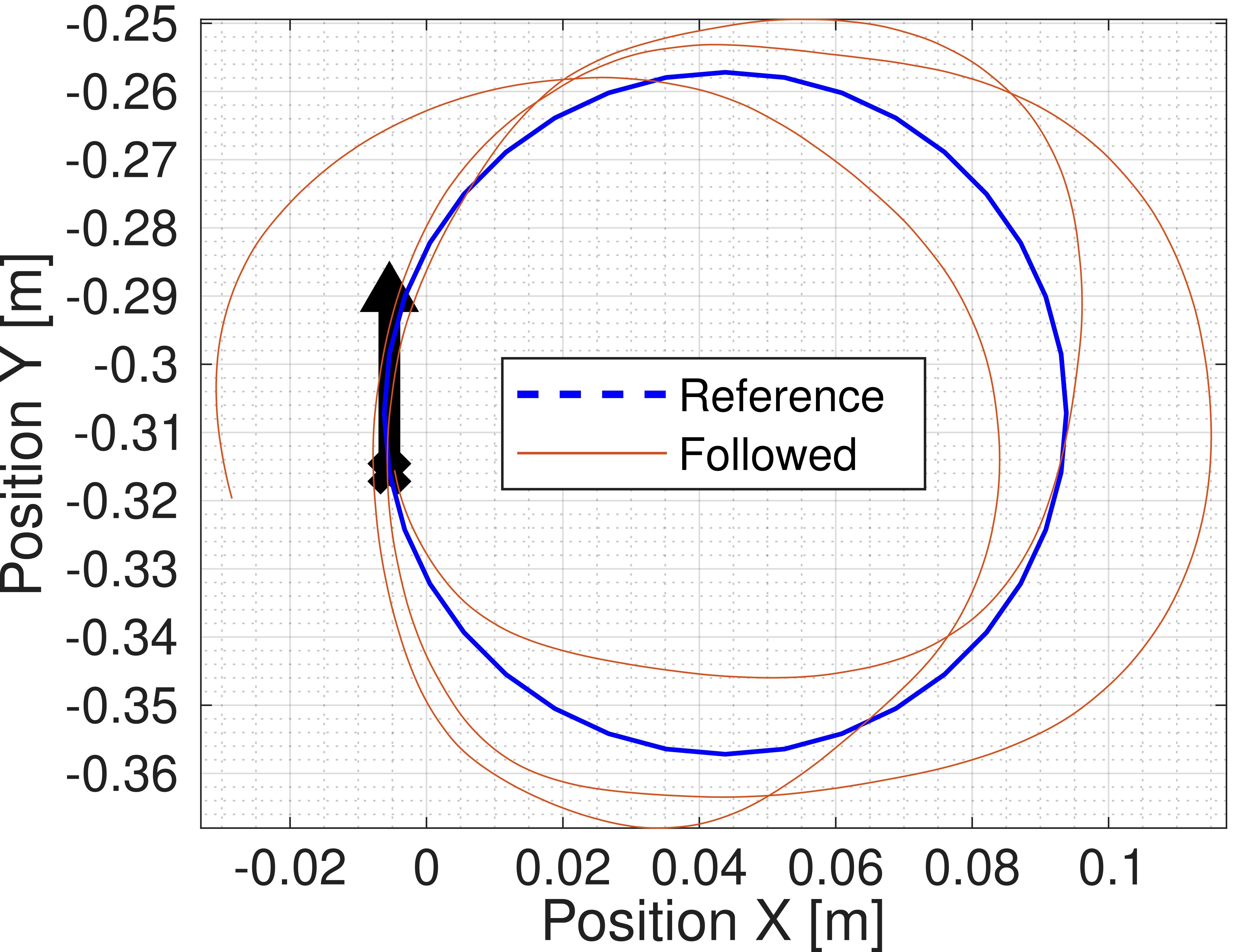}}
  \hfill
  \subfloat[\label{fig:fran:traj:NS}]{%
    \includegraphics[width=0.33\textwidth]{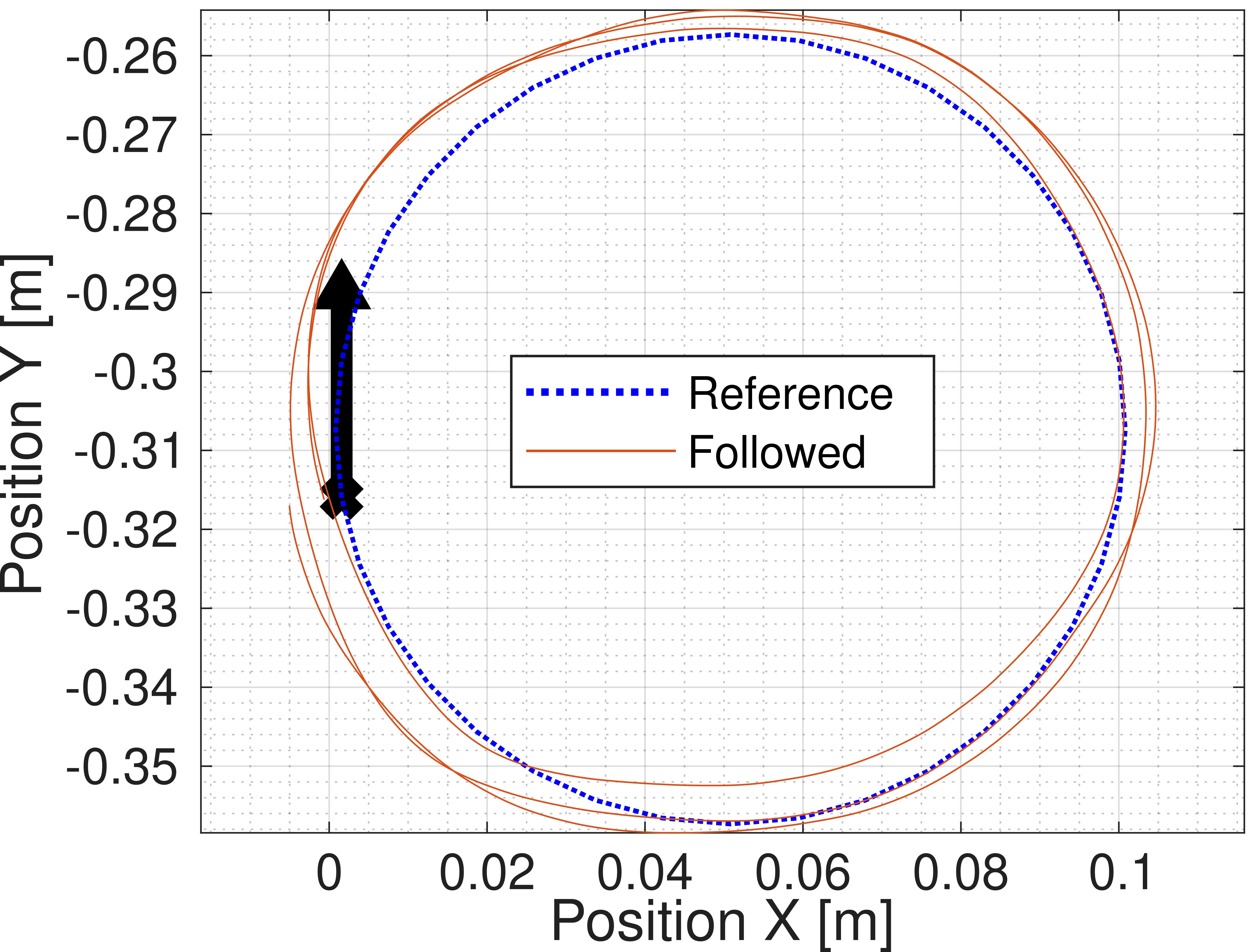}}
  \hfill
  \subfloat[\label{fig:fran:traj:NI}]{%
    \includegraphics[width=0.33\textwidth]{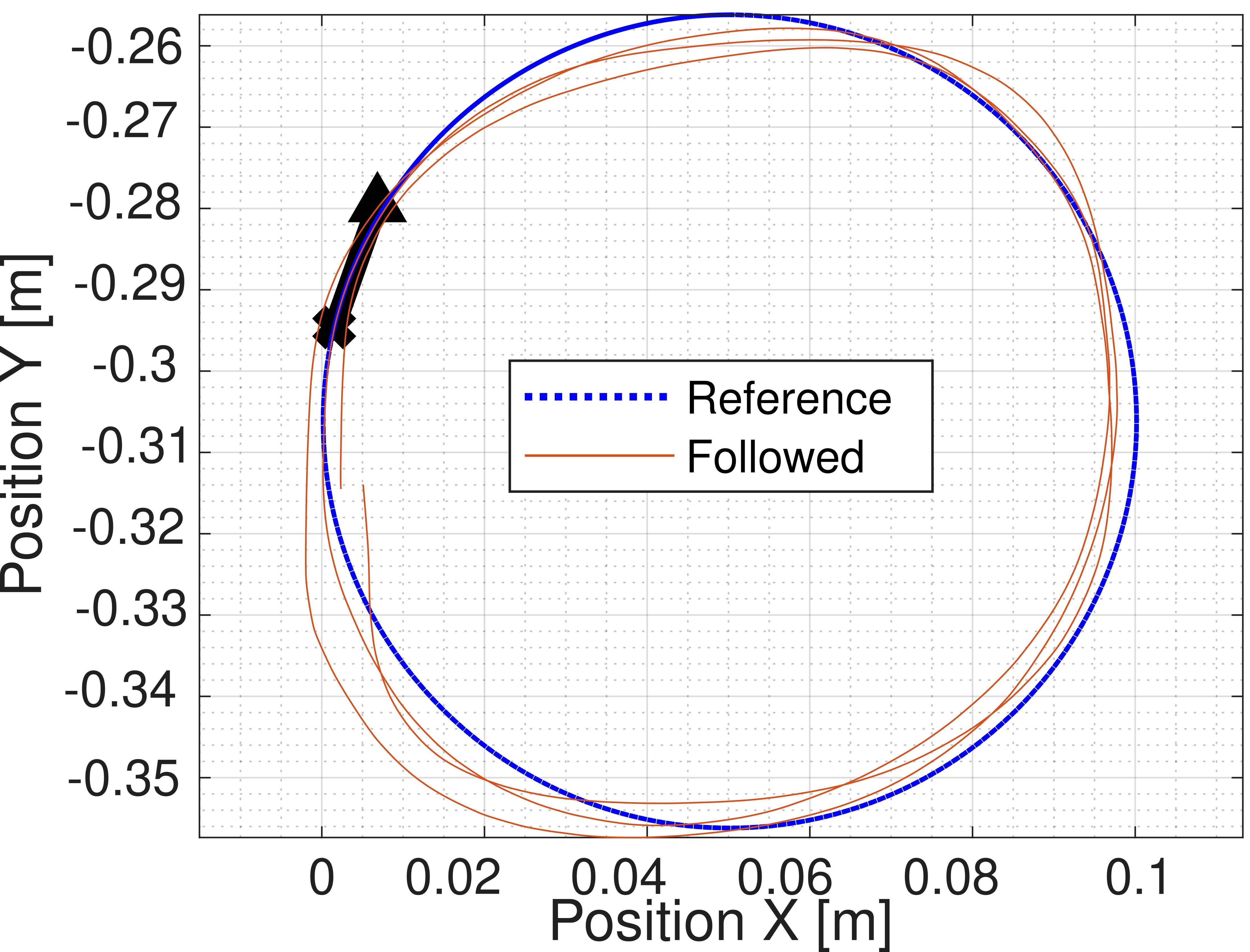}}
  \hfill
     \caption{Paths in the XY plane performed by Volunteer 08: (a) DA, (b) DS, (c) DI, (d) NA, (e) NS, (f) NI.}
  \label{fig:fran:traj} 
\end{figure*}

The parameters of the admittance force-velocity conversion were chosen as described in \cite{ReyesUquillas2021}, and were set to $\boldsymbol{M}=\textrm{diag}(1,1,1)~\textrm{kg}$ and $\boldsymbol{B}=\textrm{diag}(83.3,83.3,83.3)~\textrm{Ns/m}$. To provide a good user experience, $\boldsymbol{K}_a$ was set to the identity. 
To follow a circular path, directness and smoothness are equally important, so the weights of the performance factor are chosen to be equal, $w_1 = w_2 = 0.5$, whereas $C_1$ and $C_2$ were set to $1$ as commented in Section~\ref{subsec:eta}. The parameters of the reactive path follower were chosen to provide a good track of the path, $\lambda=1.02$ and $\rho_{min} = 0.015$~m. 

The experiments were performed following the principles of the Declaration of Helsinki in our laboratory facilities. The protocol was approved by the University of Málaga ethics committee (Protocol CEUMA 7-2023-H).
Tests were completed by ten healthy volunteers, seven male and three female; all were right-handed. Since this work does not focus on rehabilitation, volunteers were recruited from healthy people of different ages, backgrounds, and strengths. After explaining the experimental procedure, written informed consent was obtained, and a numerical ID was assigned to anonymize the data. Volunteers did not receive any training before the tests. After their arm was gripped, volunteers were asked to trace a  5 $cm$ radius circle clockwise on air with both hands \footnote{This task is inspired by Circle drawing as evaluative movement task for stroke survivors.} 
in standalone, SC mode, and in the impedance control (IC) based AAN strategy presented in~\cite{zhang_AAN} (6 tests in total). Each volunteer was asked to complete four loops to the path in each test. The first loop to the path was considered training, so it was discarded during the analysis of the results. After the experiments, volunteers were asked to complete a questionnaire. The codes used from this point on to describe every test configuration are composed as follows: the first letter corresponds to the drawing hand (\textbf{D}ominant or \textbf{N}on-dominant), whereas the second one corresponds to the control mode (stand\textbf{A}lone, \textbf{S}hared, or \textbf{I}mpedance). 
Our tests are specifically designed to check the following hypotheses: \begin{itemize}
    \item[(H1)] If performance improves in shared mode. 
    \item[(H2)] If tracing error (RMSE) decreases in shared mode. 
    \item[(H3)] If SC equalizes performance between left/right hand per participant.
    \item[(H4)] If shared mode reduces RMSE variance, it equalizes tracing precision.
    \item[(H5)] If participants are comfortable with assistance.
\end{itemize}

During standalone tests, the robot kept a grip on one of the users' forearms to acquire user information but complied with their motion, i.e., did not provide any assistance ($v_s = v_h$). In shared mode, it contributed to motion according to the proposed control law. Participants were not informed whether they were receiving assistance or not, so they could decide which test they felt more comfortable with without this knowledge. During tests, participants were provided with visual feedback on a screen, showing the circle to be traced and the current position of the robot/human limb (Figure~\ref{fig:setup}) so humans could decide whether to correct or not the trajectory and how much. To simplify the task, the path was restricted to the horizontal plane. 
Figure \ref{fig:fran:traj} shows paths completed by a random volunteer in each test configuration. Visually, it can be observed that shared control trajectories provide the best results in terms of curvature in all three loops for both the dominant and non-dominant hand. In general, SC trajectories usually present a circular shape and are more homogeneous from one iteration to the next.

\subsection{Average Test Performance}\label{sec:goalsmet}

\begin{figure*}[!thp]
  \captionsetup[subfloat]{farskip=3pt,captionskip=1pt}
    \centering
    \subfloat[\label{fig:fran:efficiency:NA}]{%
      \includegraphics[width=0.33\linewidth]{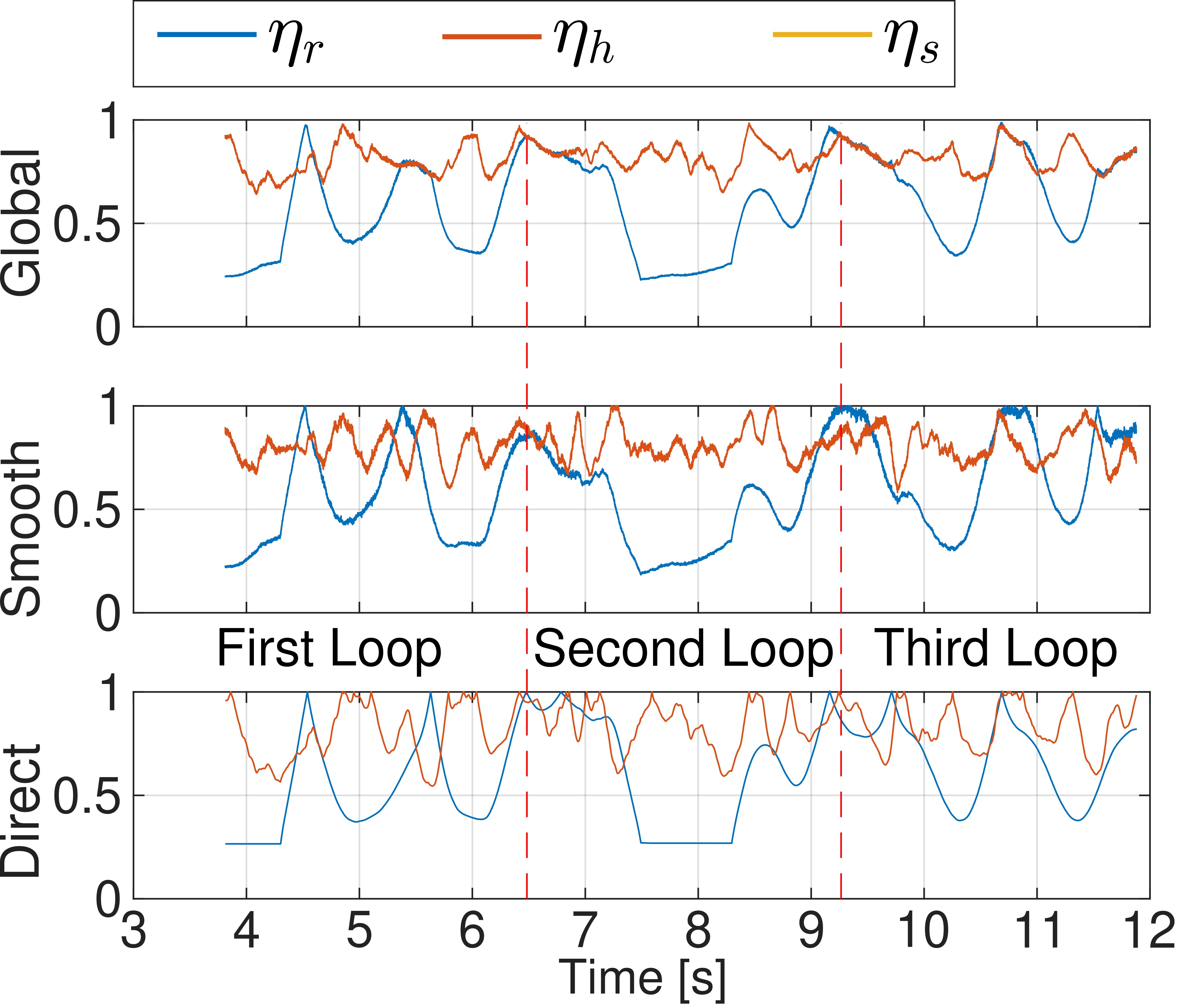}}
    \hfill
    \subfloat[\label{fig:fran:efficiency:NS}]{%
        \includegraphics[width=0.33\linewidth]{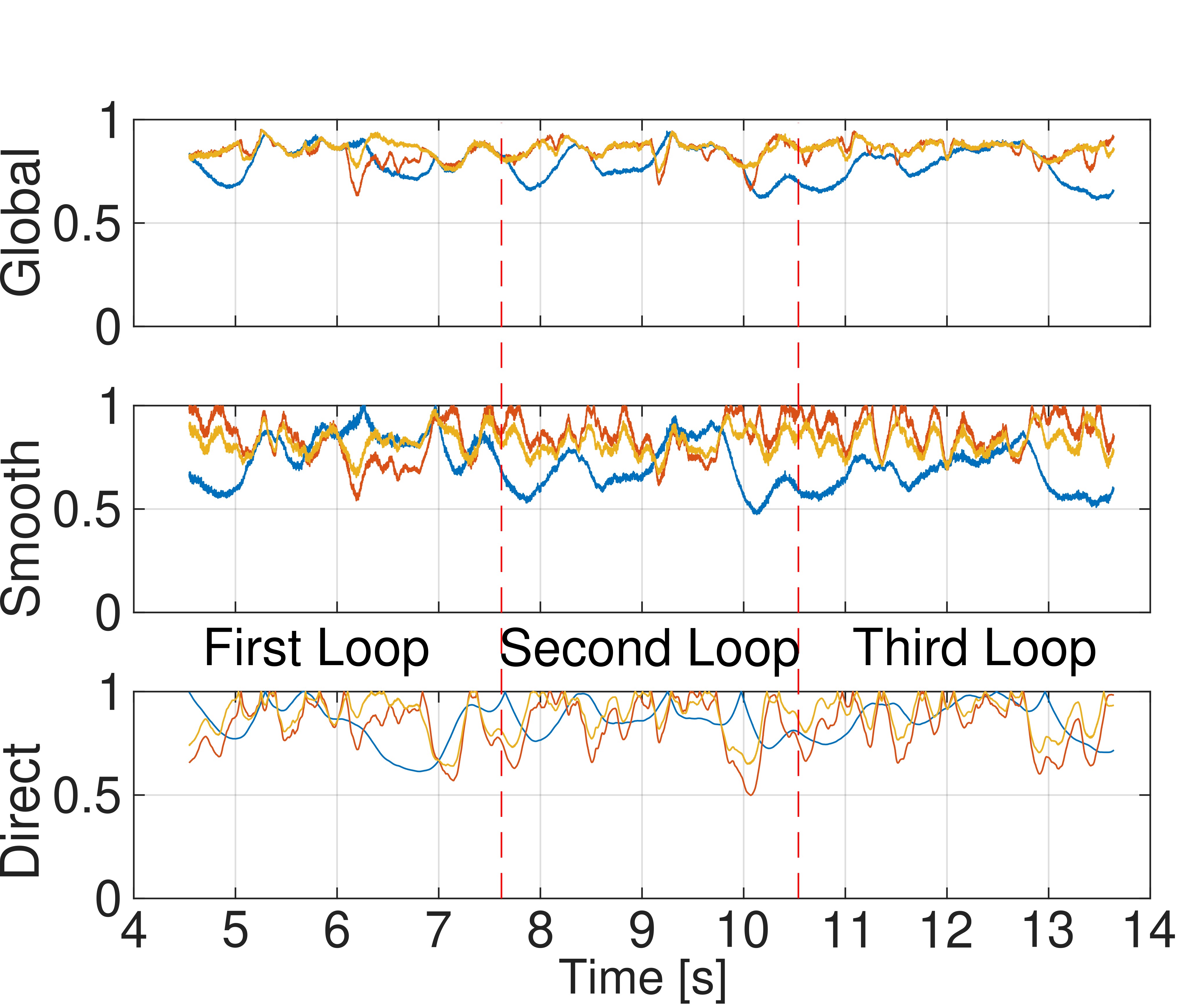}}
    \hfill  
    \subfloat[\label{fig:fran:efficiency:NI}]{%
        \includegraphics[width=0.33\linewidth]{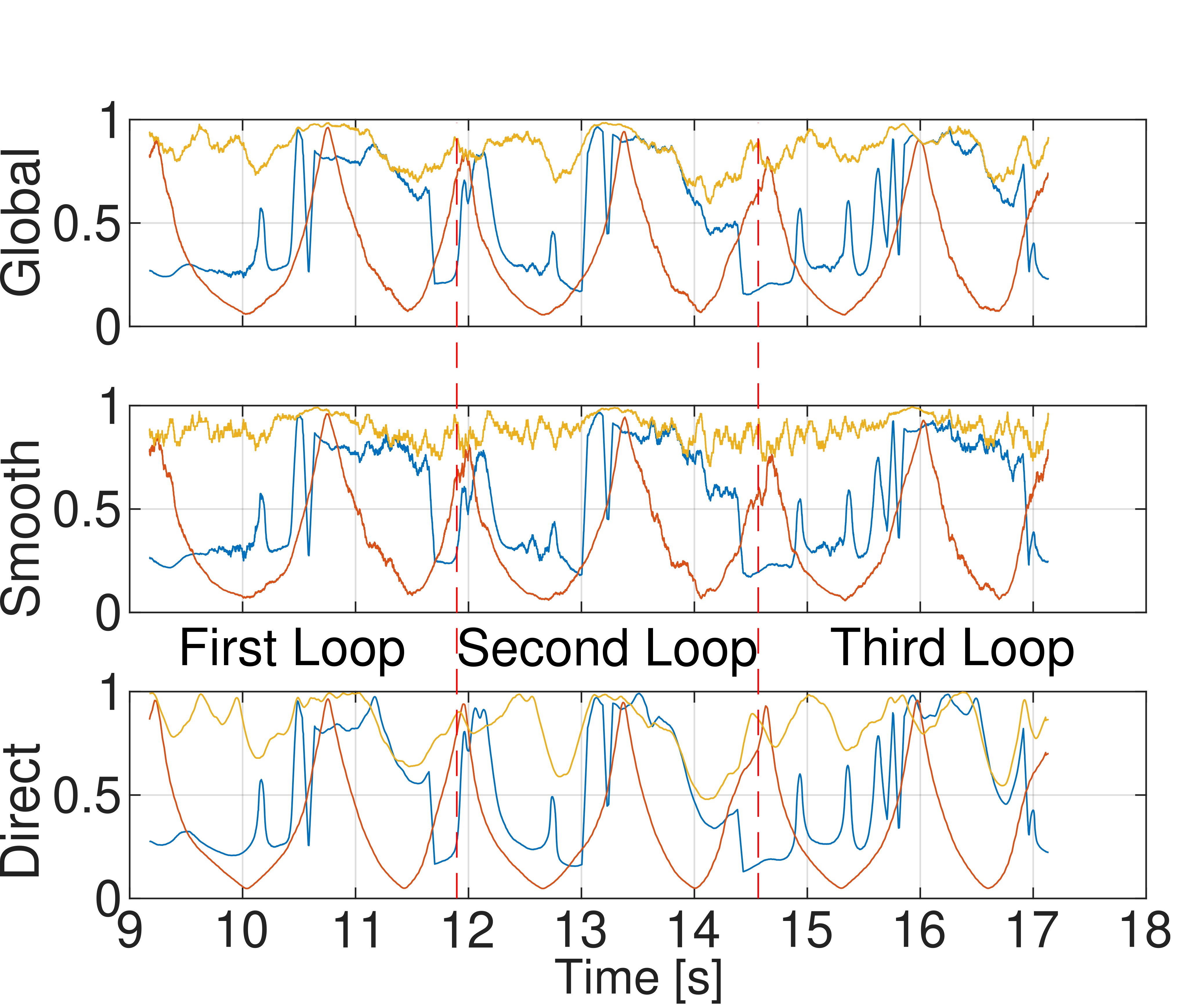}}  
    \caption{Performance for circle tracing with non-dominant hand for Volunteer 08: (a) Standalone, (b) Shared Control, (c) Impedance}
    \label{fig:fran:efficiency} 
\end{figure*} 

The proposed blend of SC relies on performance metrics to assist humans on a need basis. In this implementation, performance is measured in terms of smoothness and directness. Hence, this SC implementation is not meant to optimize strength. It must be noted that, although directness tends to make trajectories closer to reference, smoothness -and also inertia- prevents sharp changes, so, as a whole, we expect smooth, circular trajectories close to the reference but not precise tracking. To assess results, the following set of quantitative metrics is proposed. All these metrics can be obtained directly from the Franka Research 3 manipulator (no external sensors are required):
\begin{itemize}
    \item Completion time. 
    \item Forces exerted by human and robot. 
    \item Root Mean Square Percentage Error (RMSPE) between actual and reference positions (\%)~\cite{rmspe}. 
    
    \item Intervention Level (IL) (\%), deﬁned as the portion of time that volunteers contribute to motion~\cite{yanco:02}. This metric should be high, meaning that the robot and user contribute to motion simultaneously in a continuous way. As commented, the system does not move in the proposed scheme unless users provide an input command, so IL should be 100\%. However, IL is usually lower due to mechanical inertia and filtering to separate input commands from noise. Larger values mean that users provide continuous, firm input to the system.  
    \item Command Variation (\%), which measures the percentage of time with variations of more than 10\% in human forces~\cite{khoury:91}. This parameter is reportedly correlated with mental workload, deﬁned as the interaction between the demands of a task that an individual experiences and their ability to cope with these demands. It will provide a very loose indication of cognitive load, although more complex metrics would be required for a reliable study in this respect.
    \item Disagreement (\%) is the normalized angular difference between human command and emergent action~\cite{urdiales:08b}. A large disagreement usually implies opposition to assistance and, typically, to stress and frustration, i.e., the device does not respond to commands as expected.  
\end{itemize}

\begin{figure*}[!t]
      \centering
    \subfloat[Standalone mode - Force\label{fig:force:fran:NA}]{%
        \includegraphics[width=0.33\linewidth]{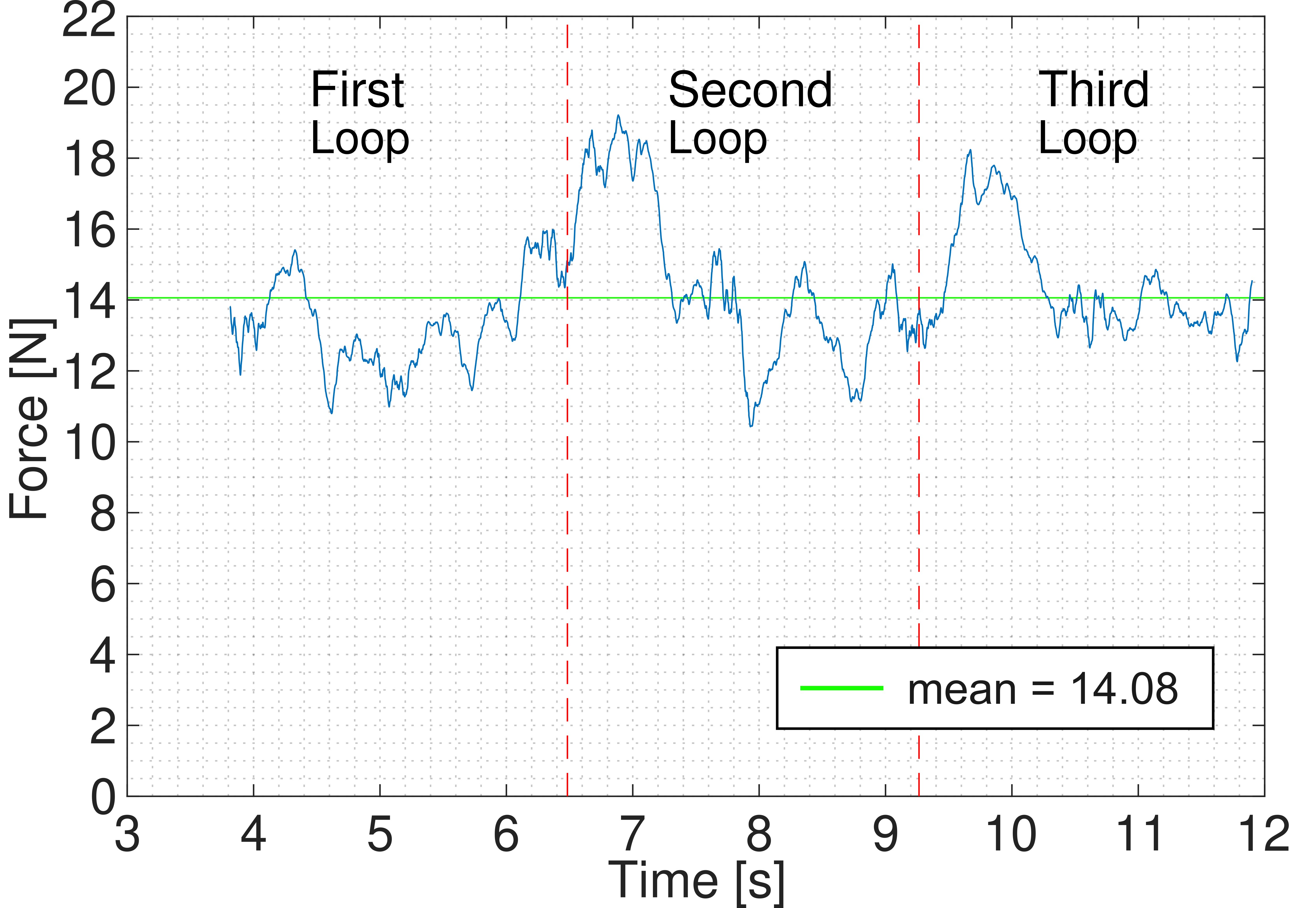}}
    \hfill
    \subfloat[Shared Control mode - Force \label{fig:force:fran:NS}]{%
        \includegraphics[width=0.33\linewidth]{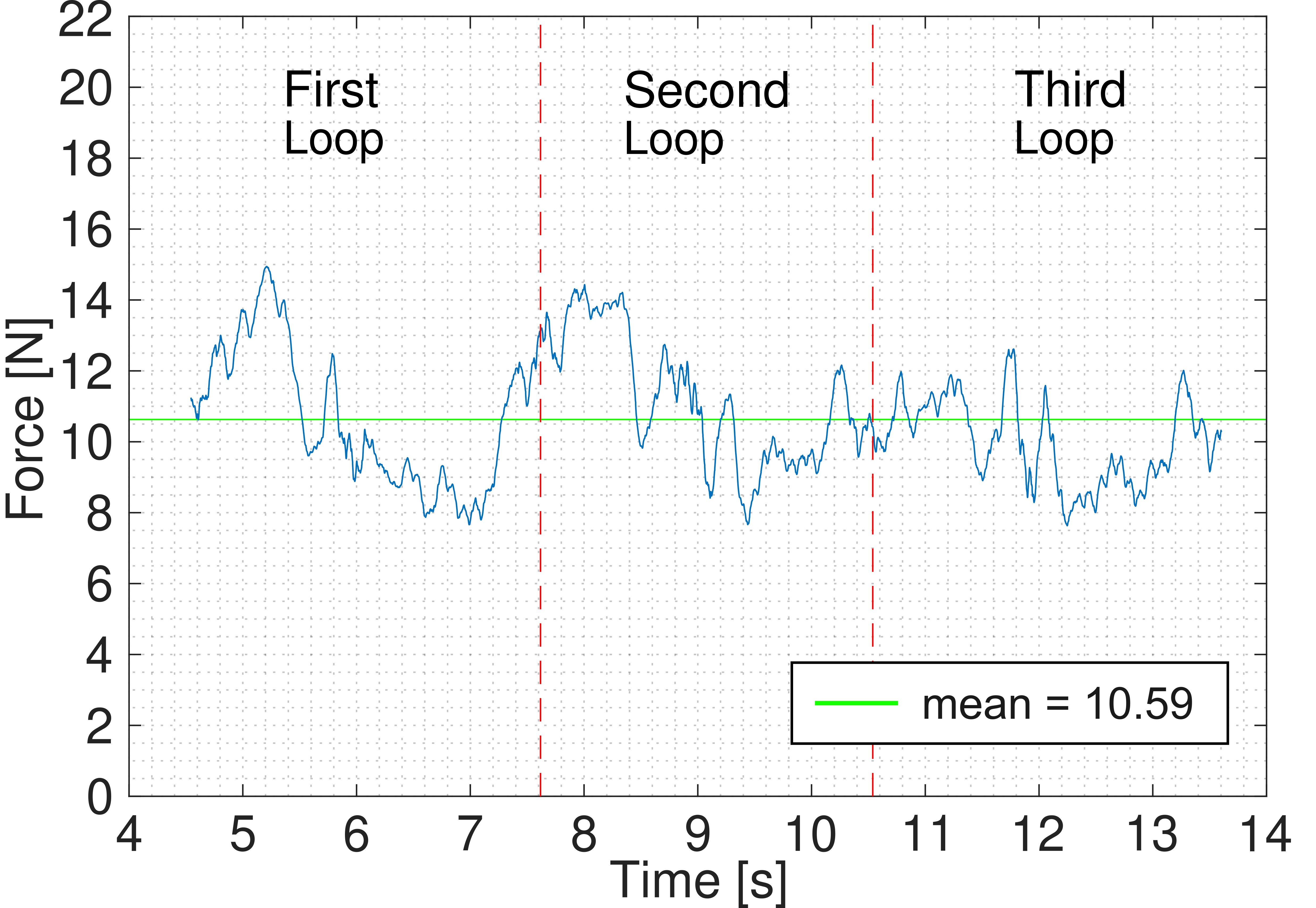}}
    \hfill
    \subfloat[Impedance mode - Force \label{fig:force:fran:NI}]{%
        \includegraphics[width=0.33\linewidth]{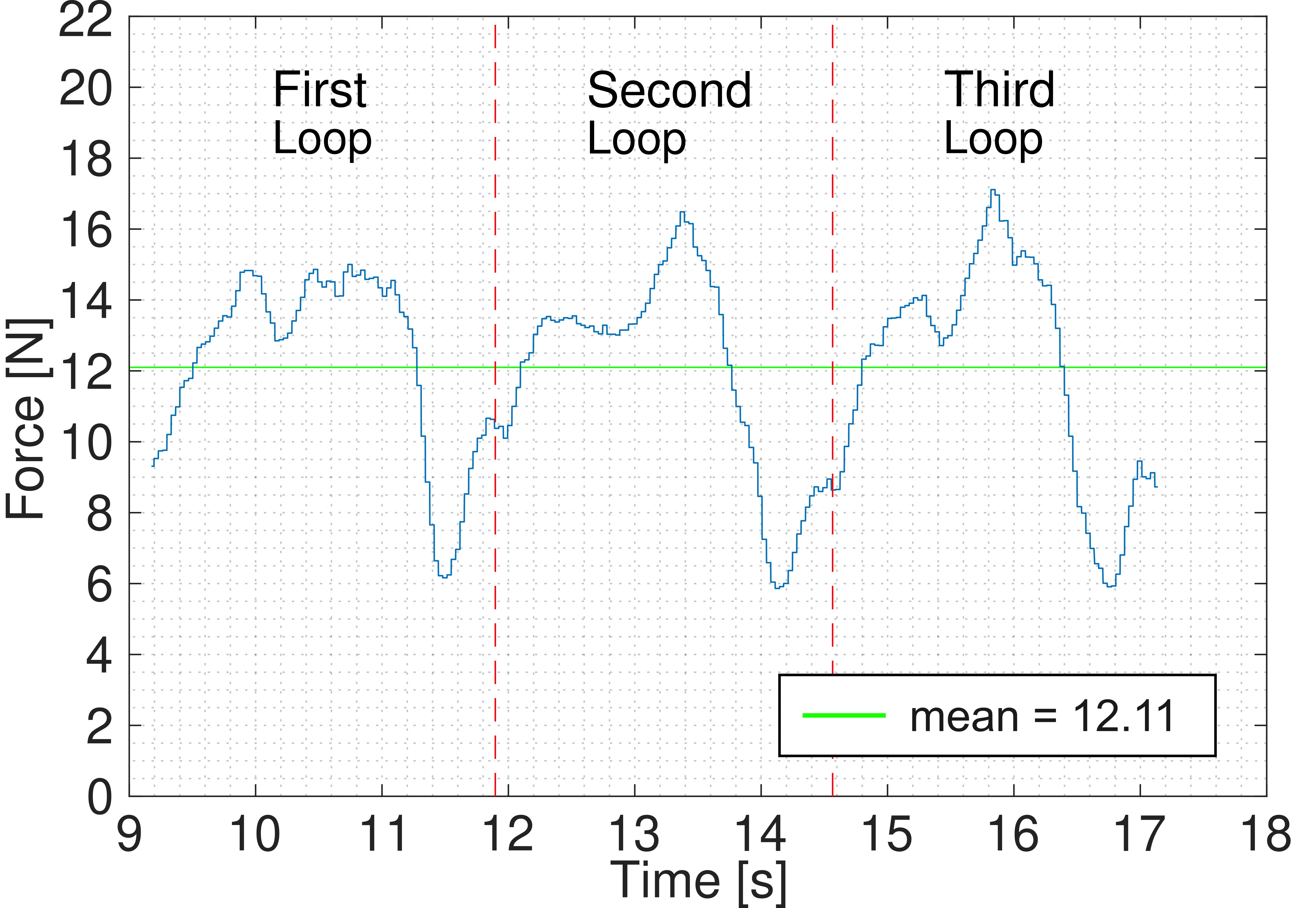}}
    \hfill \\ 
    \subfloat[Standalone mode - Disagreement \label{fig:disagr:fran:NA}]{%
        \includegraphics[width=0.33\linewidth]{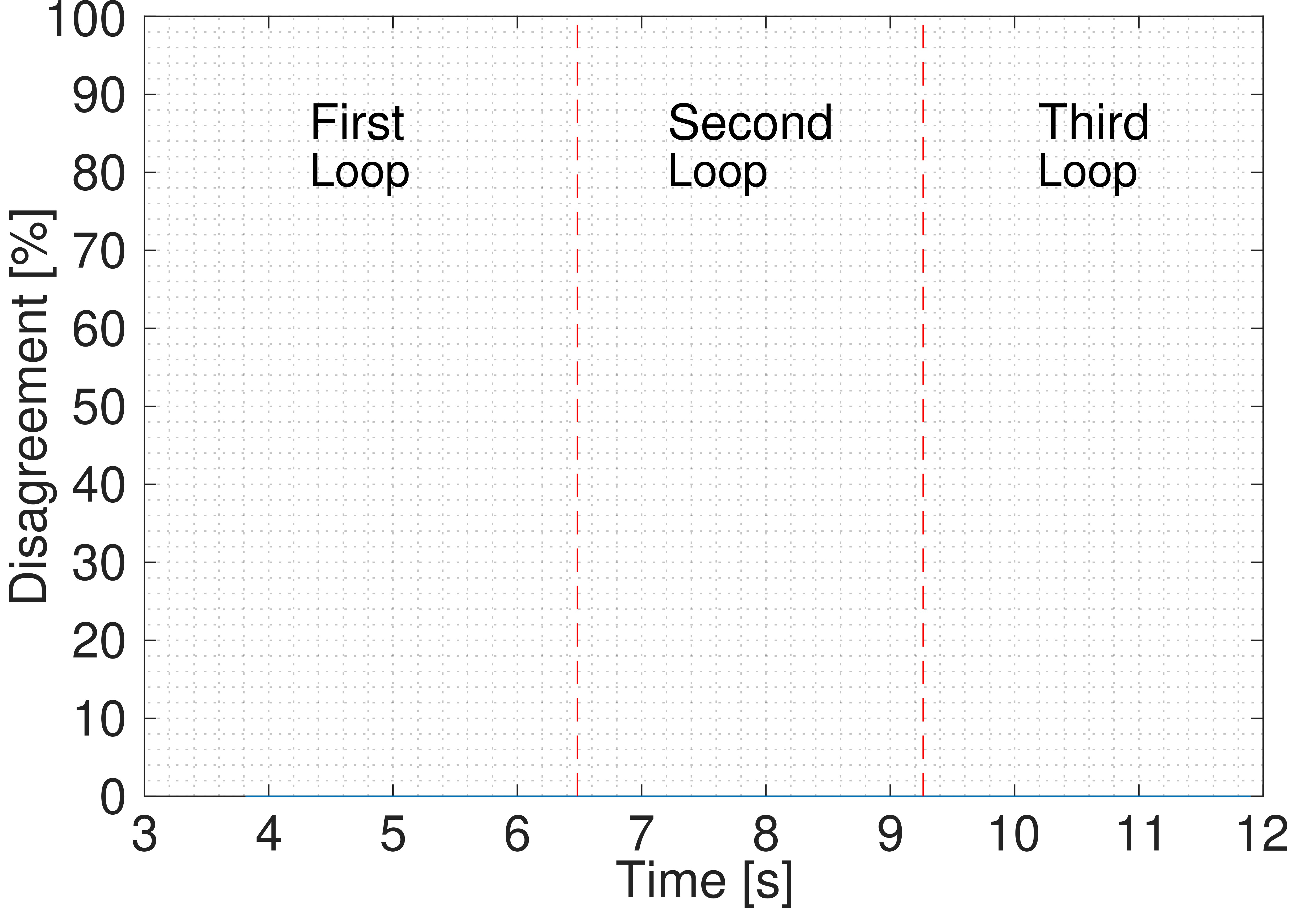}}
    \hfill
    \subfloat[Shared Control mode - Disagreement \label{fig:disagr:fran:NS}]{%
        \includegraphics[width=0.33\linewidth]{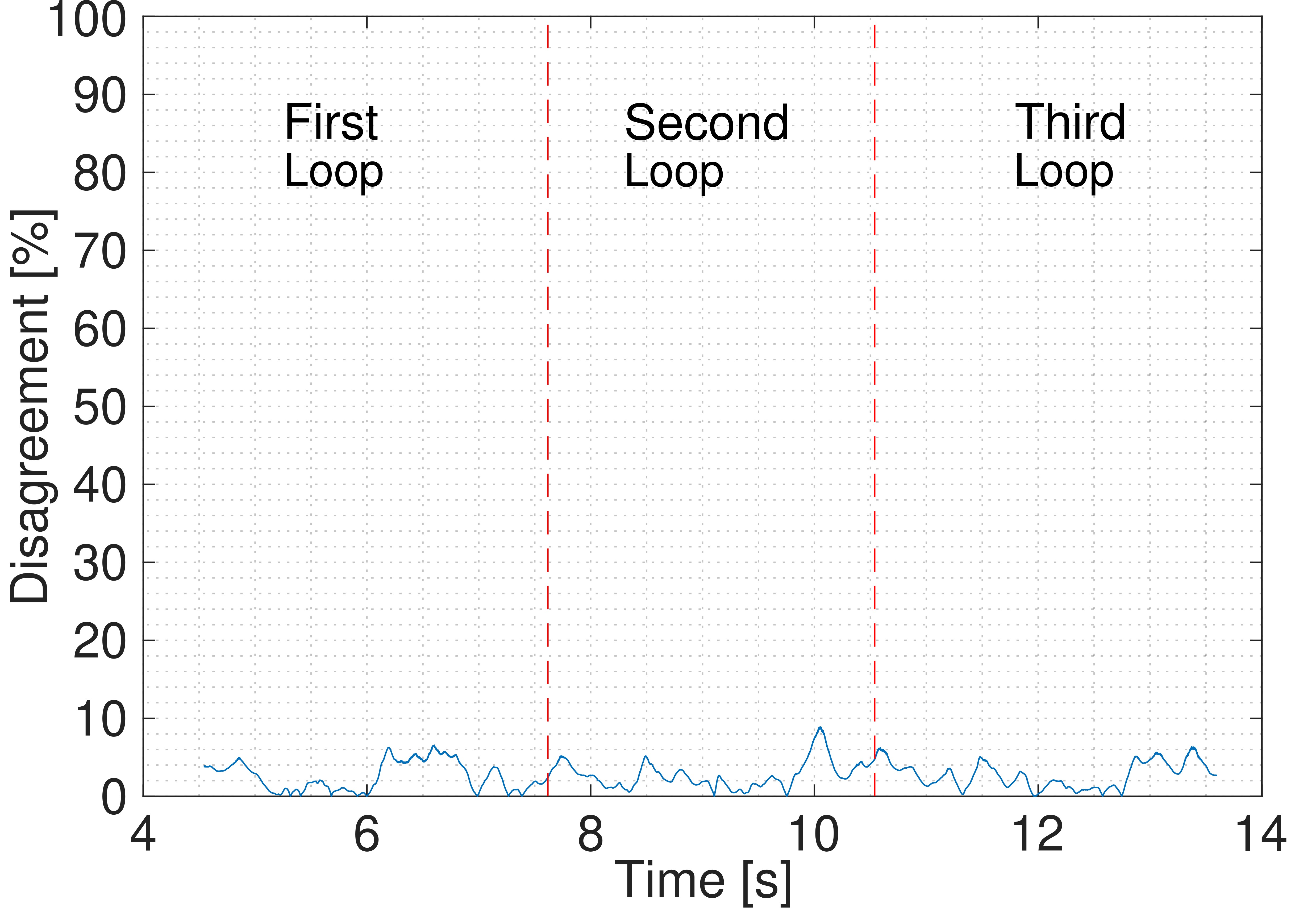}}
    \hfill
    \subfloat[Impedance mode - Disagreement \label{fig:disagr:fran:NI}]{%
        \includegraphics[width=0.33\linewidth]{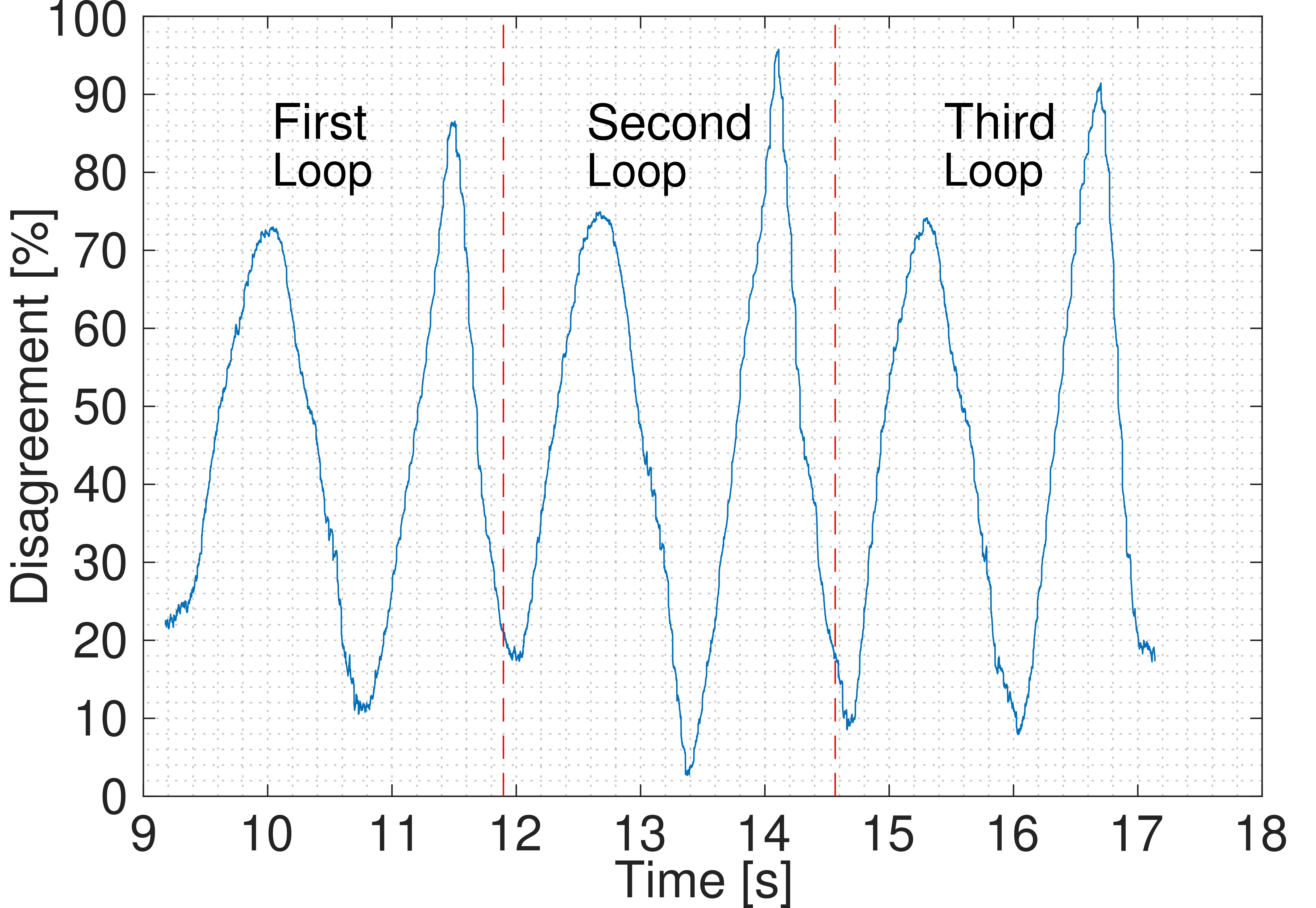}}            
    \caption{Force (top row) and Disagreement (bottom row) with non-dominant hand for Volunteer 08 in standalone (left) shared (center) and Impedance (right) modes.}
    \label{fig:force:fran} 
\end{figure*}

Additionally, local robot and human commands and their respective performance are recorded at each point of the path. Detailed data may help to understand some users' specific behaviors and compare standalone and SC modes if necessary. Figure~\ref{fig:fran:efficiency} shows an example of performances in standalone, SC mode, and IC mode for a random volunteer. In this specific example, we can observe that human performance in standalone mode -the robot still calculates a command but does not contribute to motion- presents frequent oscillations, mostly due to small path adjustments (see directness performance) and also many corrections at corners (see smoothness performance). A slight human delay with respect to robot intention can also be appreciated at corners (see smoothness performance). In shared mode, we can observe that human performance grows both in smoothness and directness, and oscillations are significantly reduced. As adjustments with the proposed control law are progressive, humans receive haptic feedback early when they produce inefficient commands and are capable of correcting them early. As a result, efficient variance is also lower. Finally, it can be noted that Impedance Control may sacrifice smoothness to boost directness, hence the performance dominant peaks. As a result, performance oscillations grow very significantly, although paths are closer to the reference circle than in standalone mode. As corrections are more local around peaks in this mode, human performance is poorer than in the proposed shared mode, even though shared control and impedance control emergent efficiencies in this case are not that different. Figure~\ref{fig:force:fran}.a-c shows the norm of the forces recorded during the same experiment. Although there are significant variations depending on the loop due to accommodation to assistance, in general, it can be observed that the norm of the force vector, on average, is lower in SC than in standalone mode and IC for this person but presents fewer fluctuations and less variation from one loop to another in impedance control. Similarly, Disagreement (Figure~\ref{fig:force:fran}.d-e) is lower in SC than in Impedance Control because, in the second case, emergent commands usually attract the gripper to the reference trajectory, despite the action the person is trying to perform. This control effect also explains the major Fluctuations in smoothness in impedance mode. Fluctuations in other modes are mostly due to inertia if users exert too little force to move the robot arm, as the robot is not expected to move in the absence of input stimuli, it stops, leading users to drastically increase strength to gain speed. Then, they let go again when motion becomes too fast for their expectations. In SC mode, the robot takes over when users underperform, as long as they provide any input, so fluctuations are significantly reduced with respect to standalone mode.

\begin{figure}[!h] 
    \centering
    \subfloat[\label{fig:eff_boxplots:global}]{%
          \includegraphics[width=0.49\columnwidth]{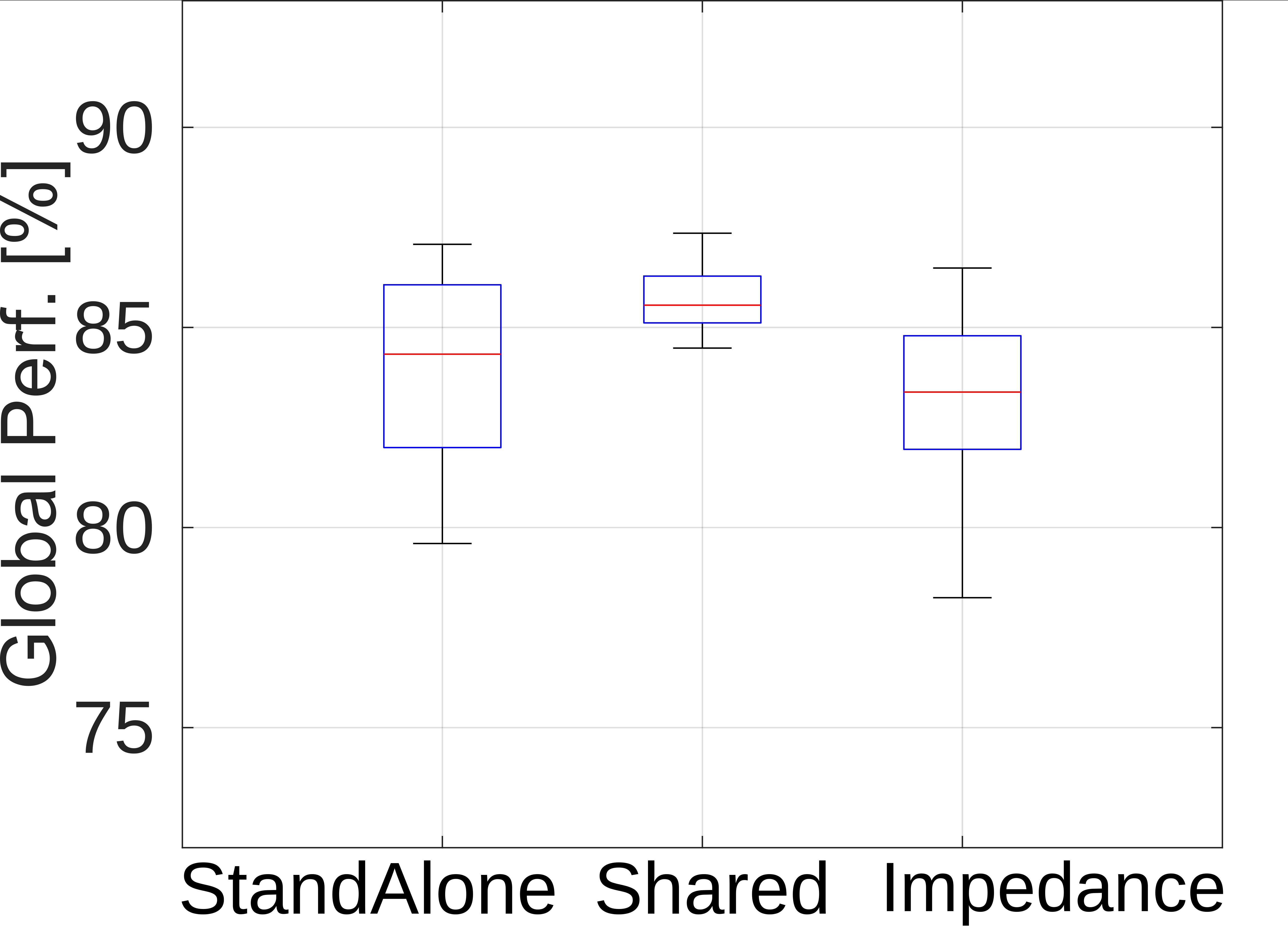}}
    \hfill
    \subfloat[\label{fig:eff_boxplots:directness}]{%
          \includegraphics[width=0.49\columnwidth]{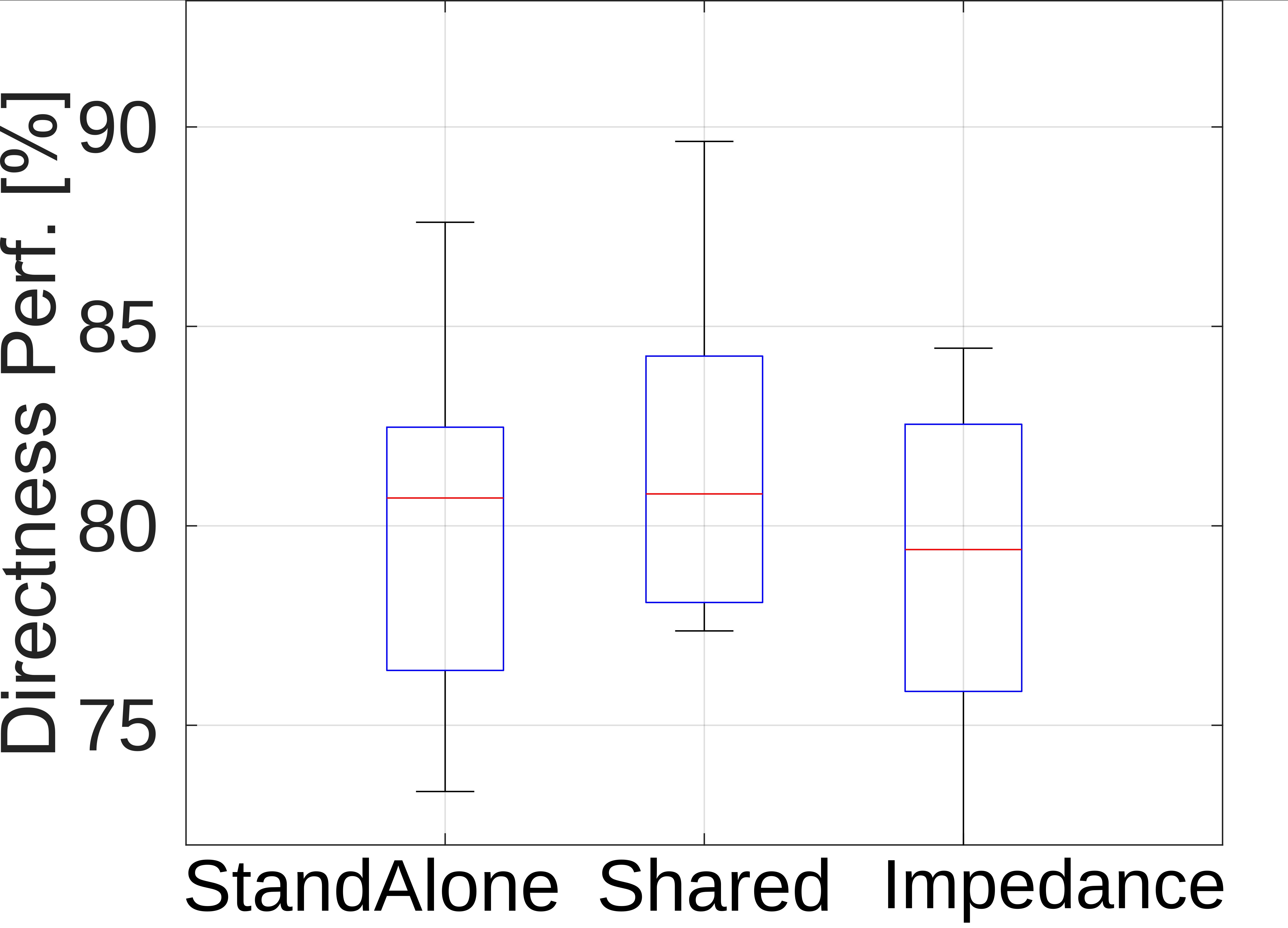}}
    \hfill
    \subfloat[\label{fig:eff_boxplots:smoothness}]{%
          \includegraphics[width=0.49\columnwidth]{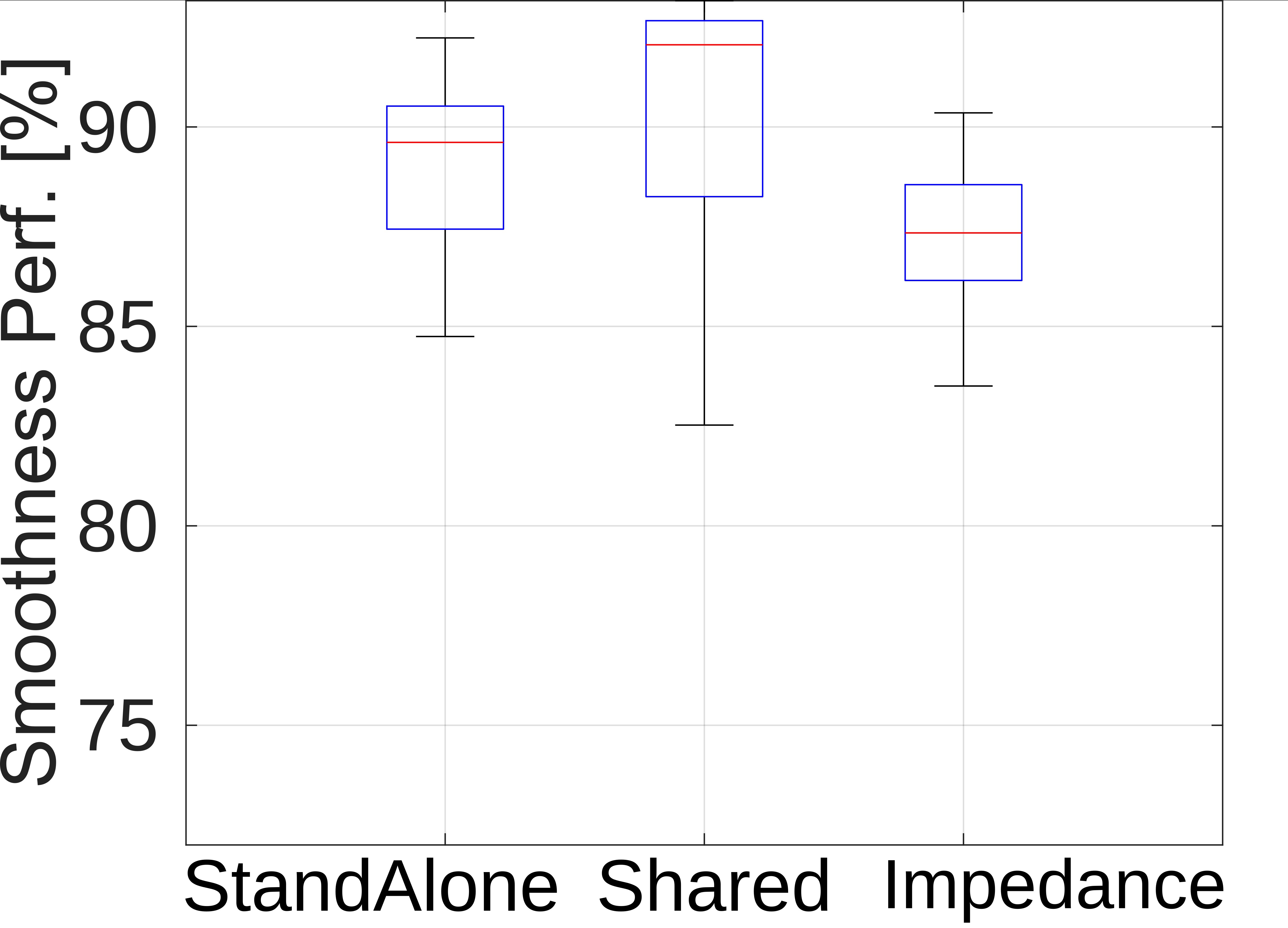}}
    \hfill
    \caption{Shared performance factors comparison.}\label{fig:eff_boxplots}
\end{figure}

\begin{figure*}[!t] 
  \captionsetup[subfloat]{farskip=3pt,captionskip=1pt}
    \centering
    \hspace{1cm} \hfill
    \subfloat[\label{fig:boxplots:force}]{%
          \includegraphics[width=0.25\linewidth]{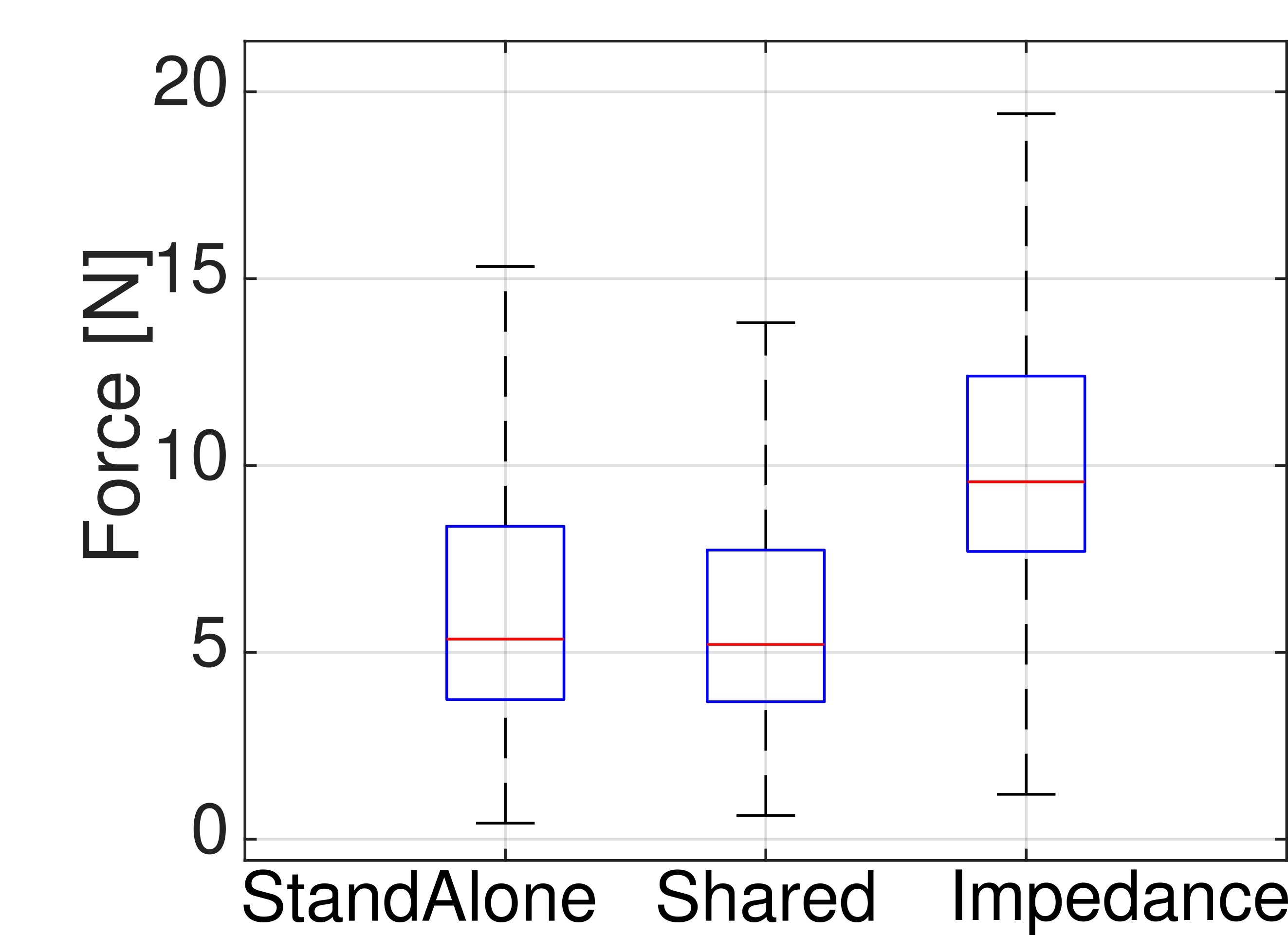}}
    \hfill
    \subfloat[\label{fig:boxplots:rmspe}]{%
          \includegraphics[width=0.25\linewidth]{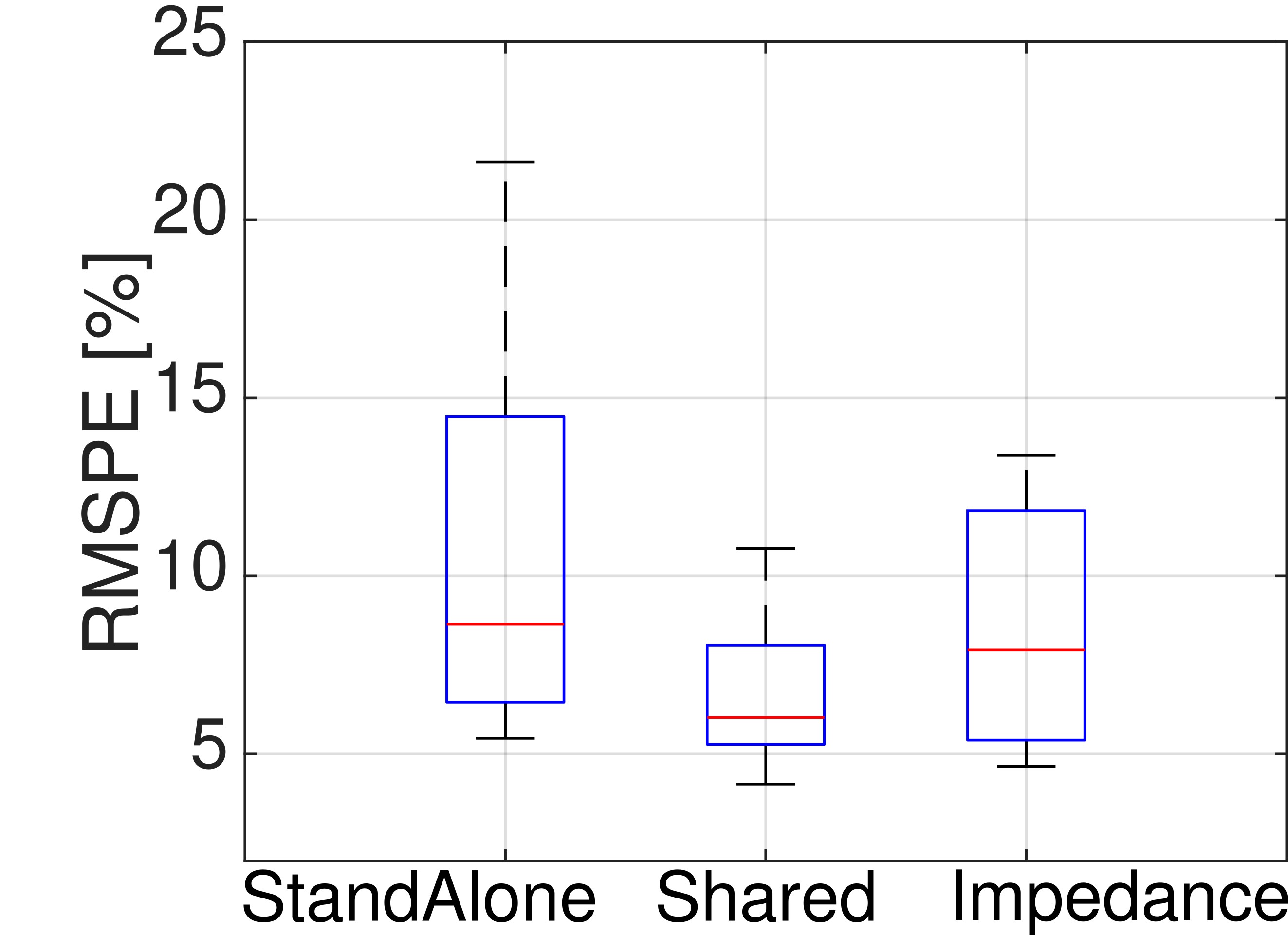}}
    \hfill
    \subfloat[\label{fig:boxplots:disagreement}]{%
         \includegraphics[width=0.25\linewidth]{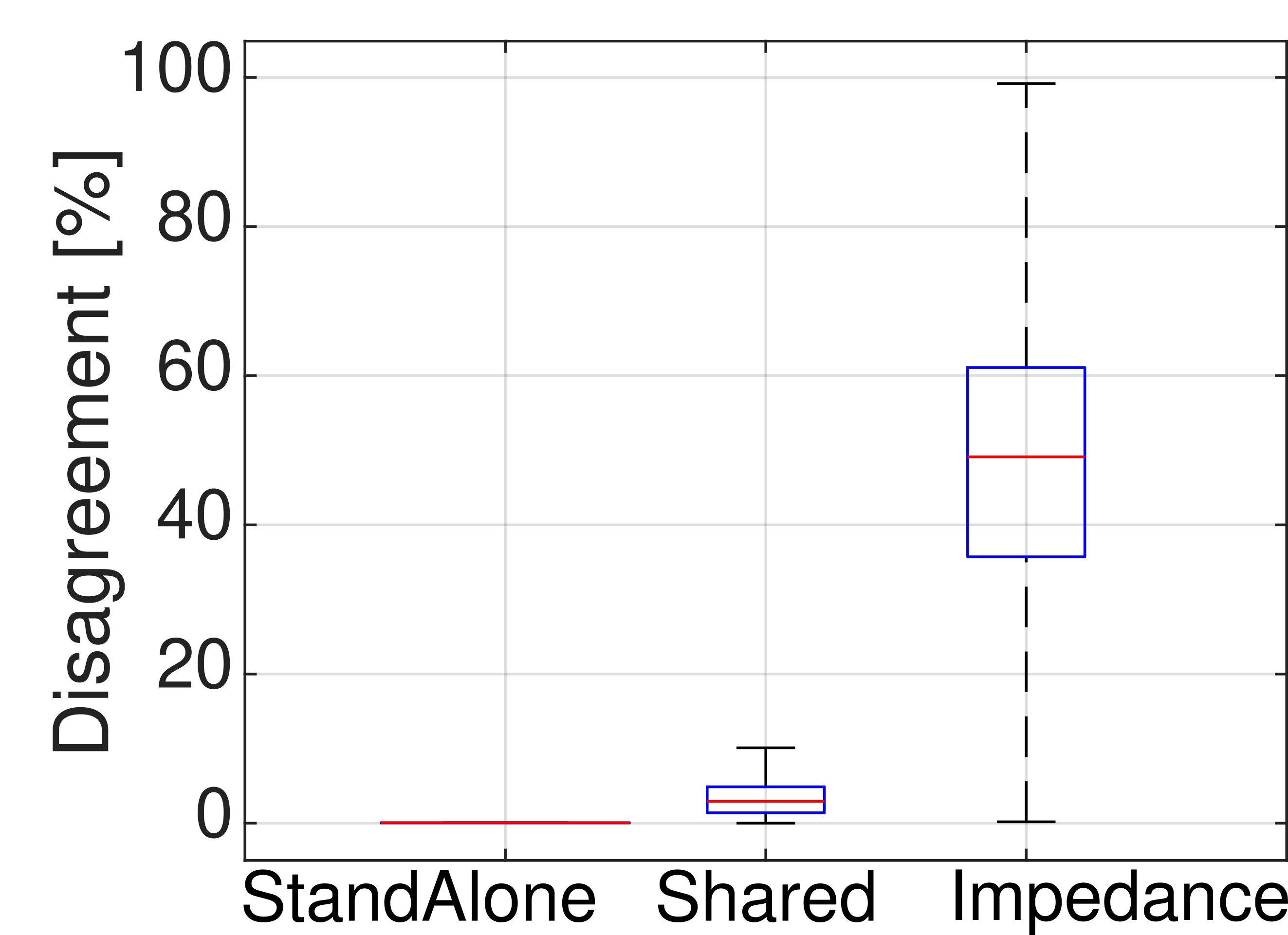}}
    \hfill 
    \hspace{1cm} \hfill \\ 
    \hspace{1cm} \hfill
    \subfloat[\label{fig:boxplots:intervention}]{%
          \includegraphics[width=0.25\linewidth]{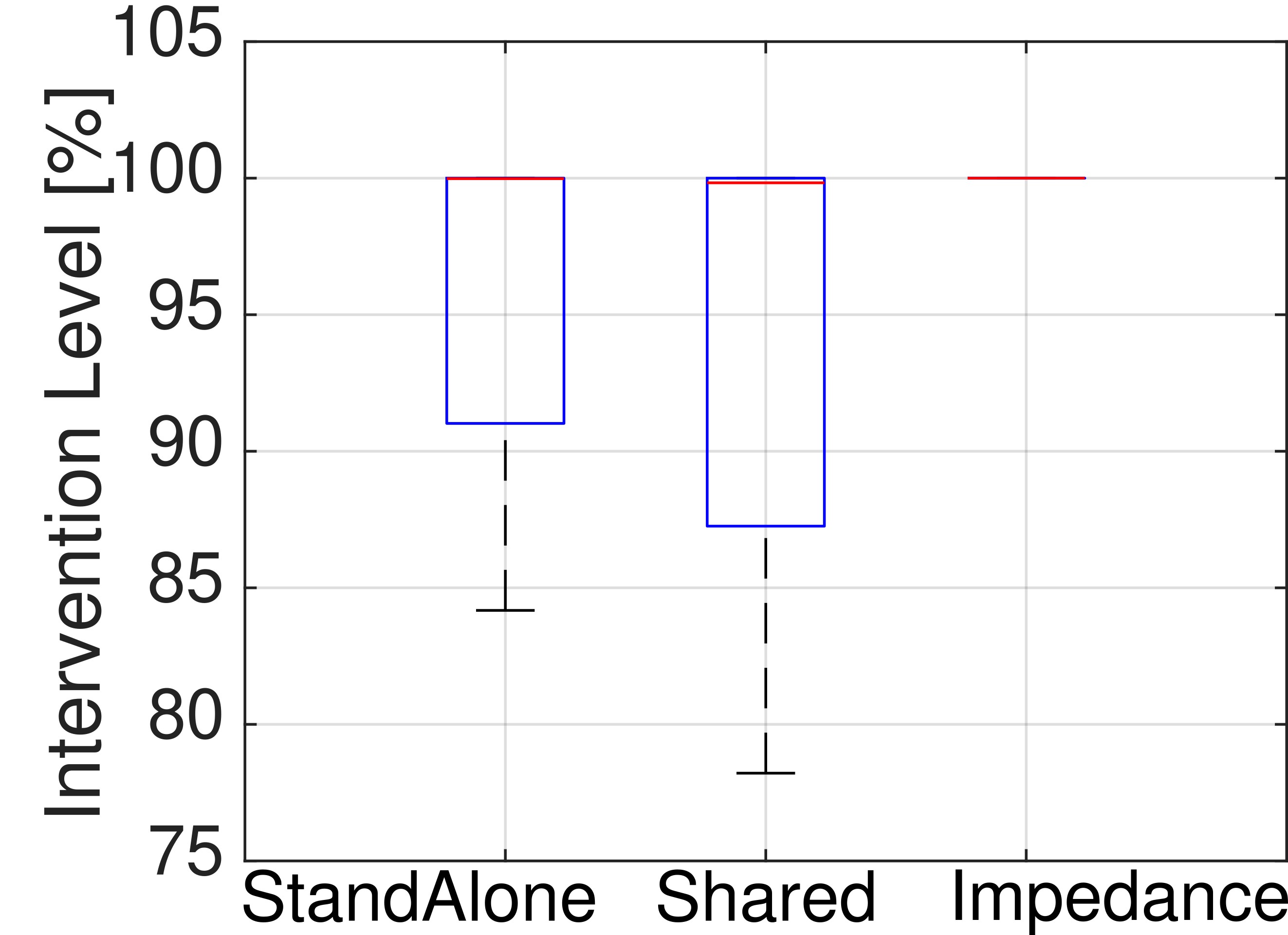}}
    \hfill 
   \subfloat[\label{fig:boxplots:joy_var}]{%
          \includegraphics[width=0.25\linewidth]{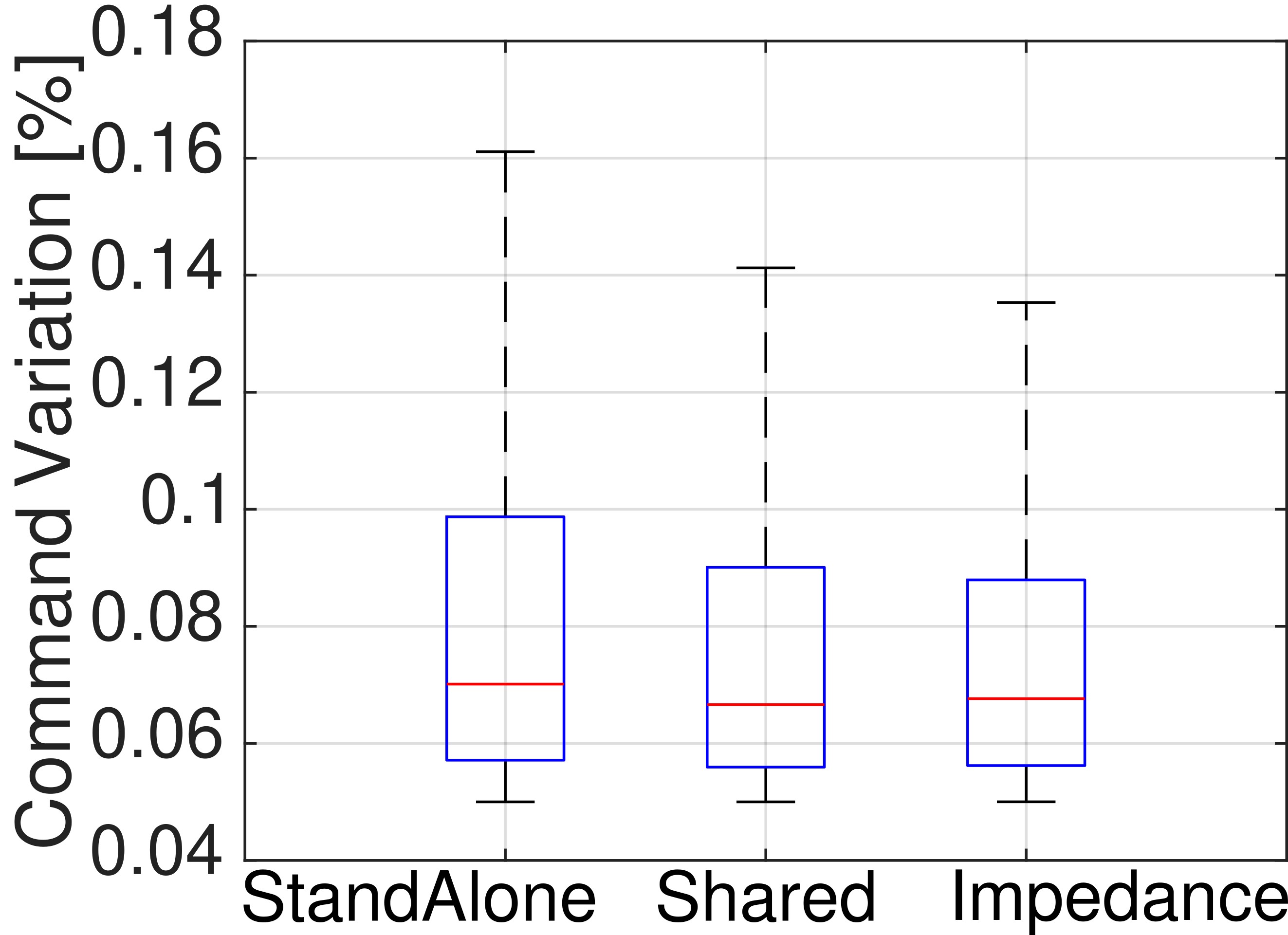}}
    \hfill
    \subfloat[\label{fig:boxplots:compl_time}]{%
          \includegraphics[width=0.25\linewidth]{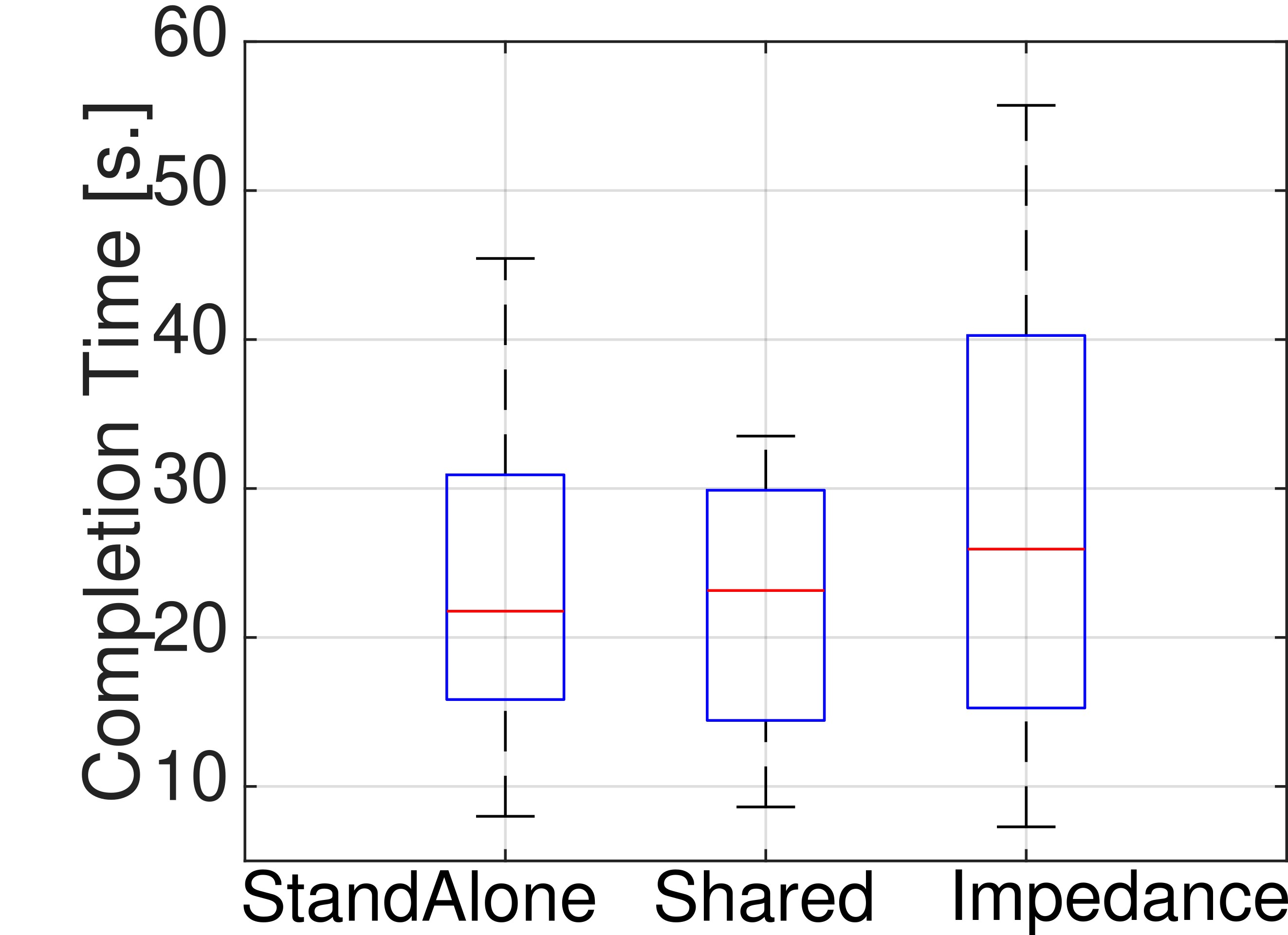}}
    \hfill \hspace{1cm} 
    \caption{Average tests performance.}\label{fig:boxplots}
  \end{figure*}

\subsection{Result Analysis}

Figure~\ref{fig:eff_boxplots} shows the performance of our 10 participants in tests in all three control modes with dominant and non-dominant hands, i.e., 60 tests. As commented, shared control is not designed to boost performance as a rule but to raise human performance when needed. Hence, rather than a large increase in mean performance, a decrease in variance was expected. This means participants offer a more homogeneous performance, even operating with their non-dominant hands. 

Nevertheless, it can be observed that the distribution median is slightly higher in SC mode with respect to the standalone mode. Focusing on performance factors instead, we can observe that directness does not improve much in shared mode, although the distribution is better. It is interesting to note that IC, which favors directness over other considerations, does not outperform the other control modes in this respect. In general, volunteers in this test were people who performed well on their own. 

Finally, it can be observed that smoothness improves significantly in SC mode as a side effect of its control law. Improvement in global performance and smoothness were validated by ANOVA both for shared vs standalone mode (p-values of 0.0319 and $<0.01$, respectively) and SC vs IC (p-values of $<0.01$ and 0.025, respectively).

Figure \ref{fig:boxplots}.a-f show remaining metrics for every participant in our tests (10 participants in total) in all three control modes (Standalone, SC and IC) with dominant and non-dominant hands, i.e. 60 tests. 
Average forces (Figure \ref{fig:boxplots}.a) are similar in standalone and SC modes, but forces clearly increase in Impedance Mode. This was to be expected because, in this mode, people often tend to counteract robot corrections. This effect can also be appreciated in Disagreement (see Figure \ref{fig:boxplots}.c). It can be observed that human commands often match robot commands, probably due to the progressive nature of corrections and the increase in smoothness.

Figure \ref{fig:boxplots}.b shows RMSPE, measured as the difference of the figure traced by users with respect to the reference on screen. A clear decrease in average error can be observed in SC mode. Although the proposed control law has not been designed to optimize RMSPE for any specific path, SC increases impedance progressively when users move away from desired trajectories. This haptic feedback usually avoids larger errors. IC also reduces RMSPE as a whole, but not as much as SC, and error variance is higher. This result supports the performance homogenization effect of SC.

Regarding Intervention Level, which, as reported, concerns human input, it can be observed in Figure \ref{fig:boxplots}.d that, in average, it is close to 100\% in all cases. This was to be expected, because the system is not supposed to move unless a human command is provided, except for inertia. It is interesting to note, though, that in Standalone and SC mode, human commands present some variation, whereas in IC mode humans contribute to control all the time. Lower IL values do not correspond to stops, as proven by shorted completion time averages in these modes (Fig. \ref{fig:boxplots}.f) with respect to IC. Larger average force values in IC (\ref{fig:boxplots}.a) seems to point out that users struggle more with the robot in IC mode, hence their continuous intervention, whereas in other modes, particularly in SC, they agree more with ongoing motion.

In SC mode, the robot is continuously affecting motion, so people tend to interact more with the system, as supported by the reported force variation increase. Despite this increased interaction, it can be observed that Command Variation (Figure \ref{fig:boxplots}.e) presents no significant differences in average and, furthermore, variance is lower in SC mode, pointing out that, as far as our indicators go, the cognitive load does not seem to increase in average and is more homogeneous for the different users and situations. In this case, IC offers very similar results.

Completion times present no significant differences in these tests (Figure \ref{fig:boxplots}.f) from standalone to shared mode, although there is a variation decrease, as usual. In IC mode, completion time significantly increases in variation. This reflects some human/robot struggle to keep in the right direction (e.g., to modify curvature or to move closer to the reference figure).

\begin{figure*}[t] 
  \captionsetup[subfloat]{farskip=3pt,captionskip=1pt}
    \centering
    \subfloat[\label{fig:scatterplots:aut:force-eff}]{%
          \includegraphics[width=0.33\linewidth]{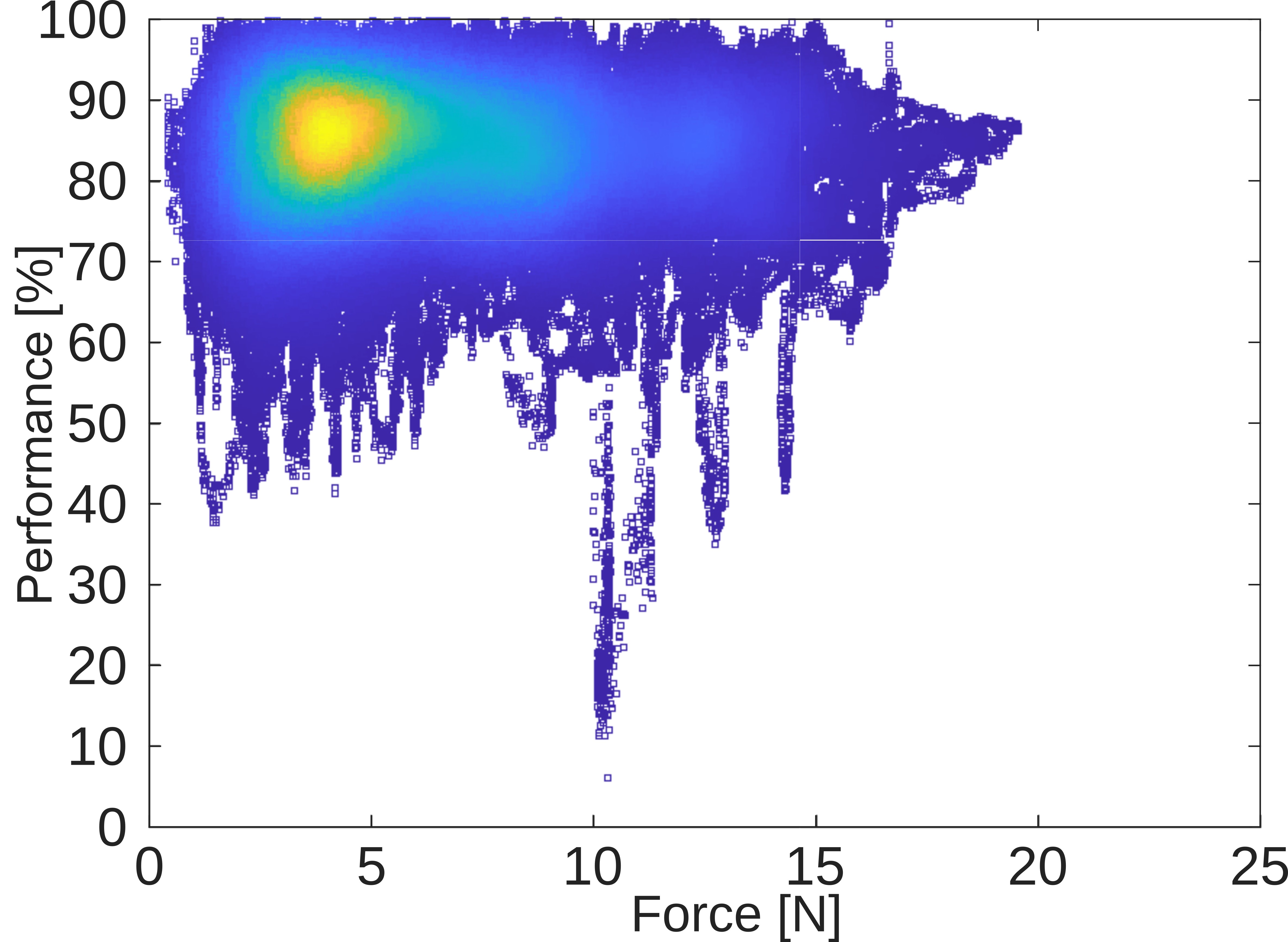}}
    \hfill
    \subfloat[\label{fig:scatterplots:aut:force-dis}]{%
          \includegraphics[width=0.33\linewidth]{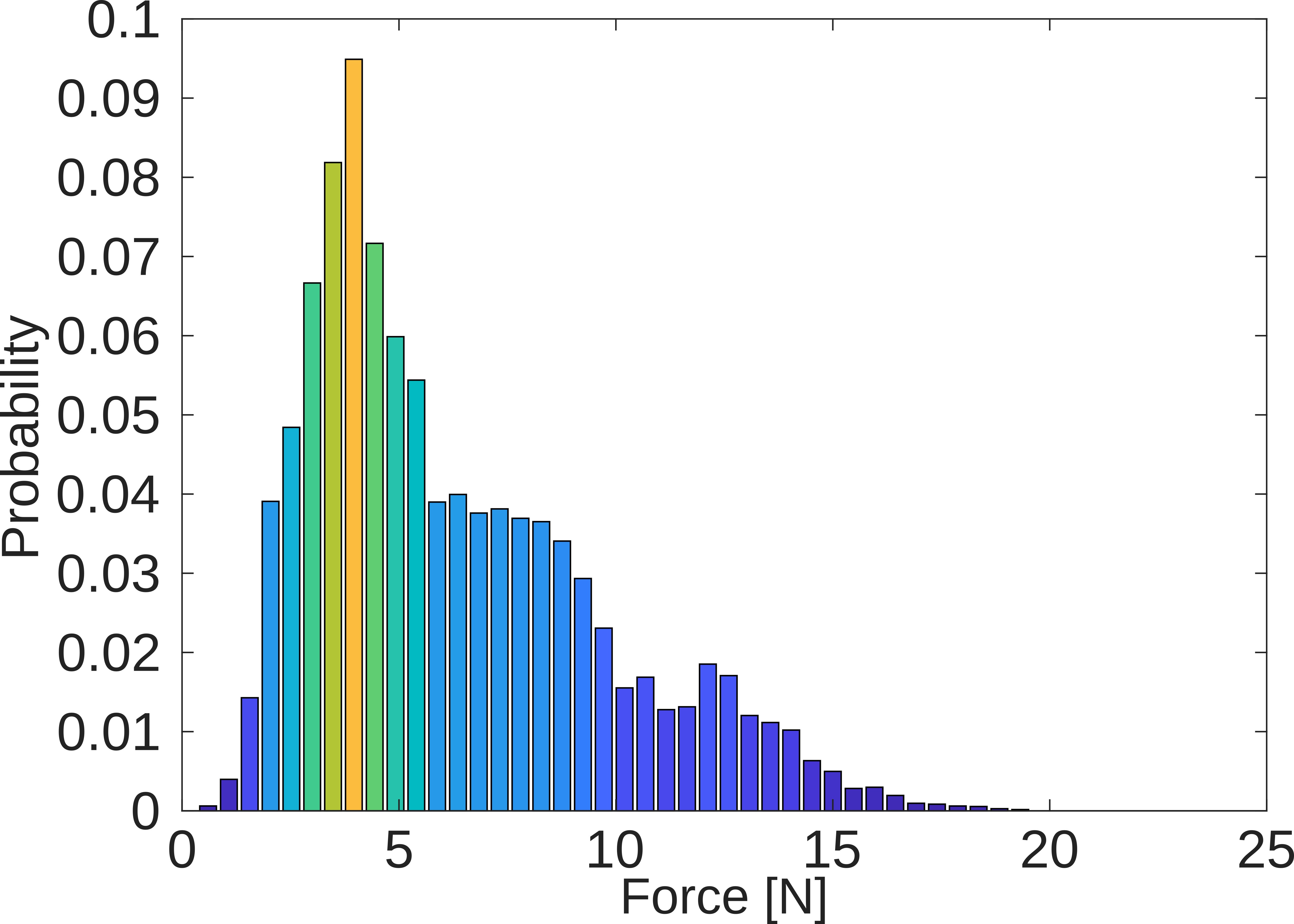}}
    \hfill
    \subfloat[\label{fig:scatterplots:aut:dis-eff}]{%
          \includegraphics[width=0.33\linewidth]{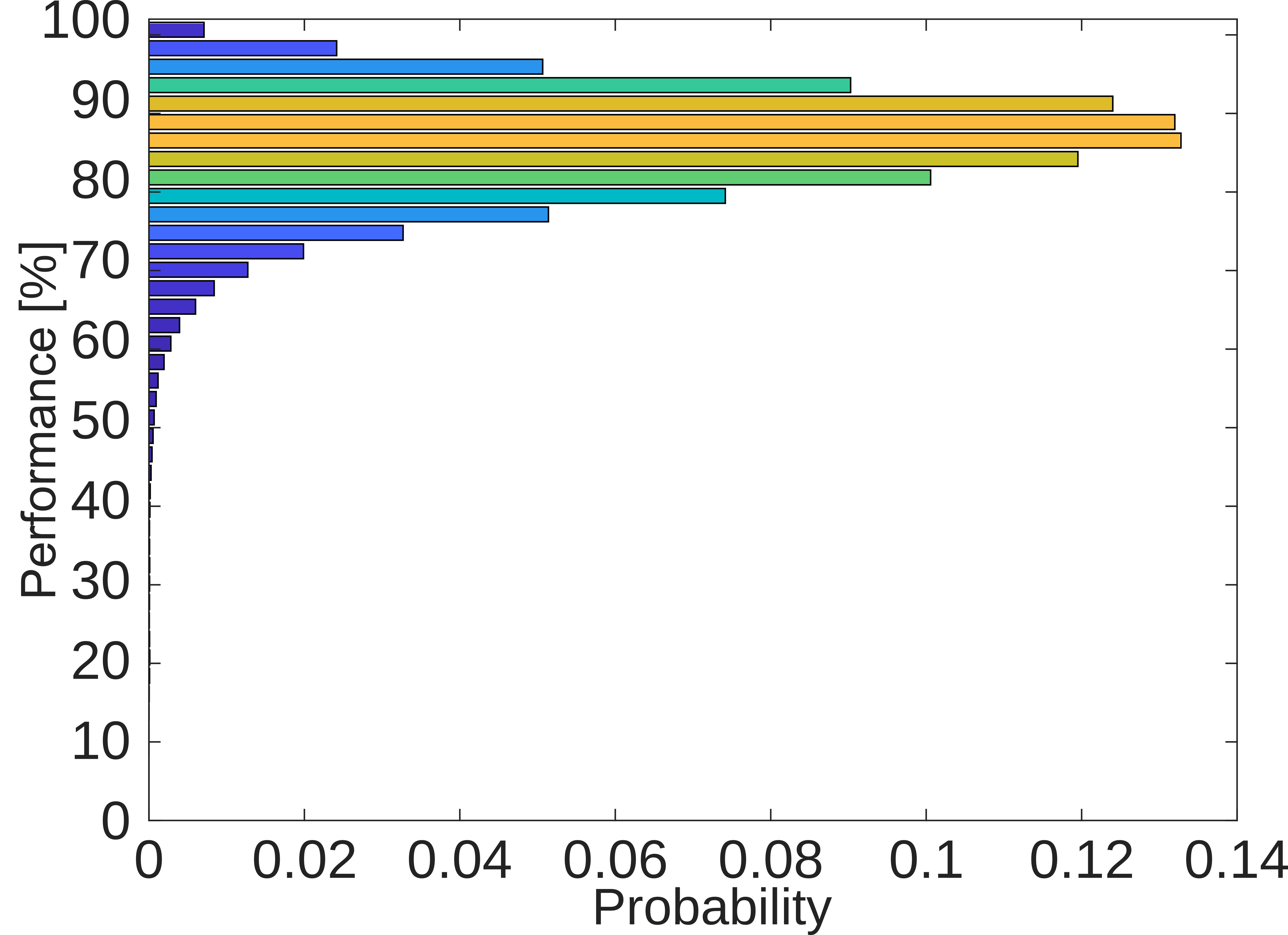}}
    \hfill \\
    \subfloat[\label{fig:scatterplots:shared:force-eff}]{%
          \includegraphics[width=0.33\linewidth]{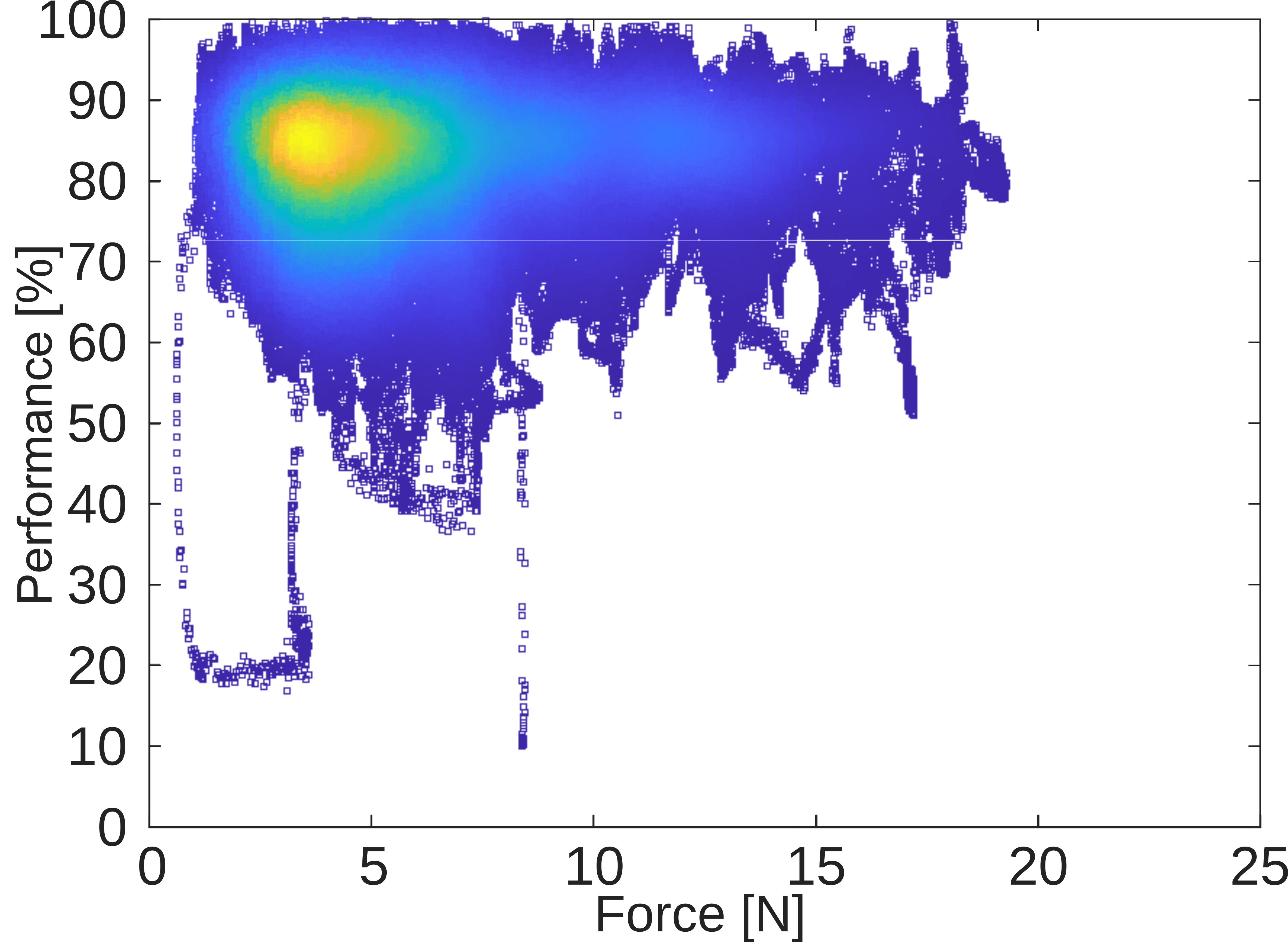}}
    \hfill
    \subfloat[\label{fig:scatterplots:shared:force-dis}]{%
          \includegraphics[width=0.33\linewidth]{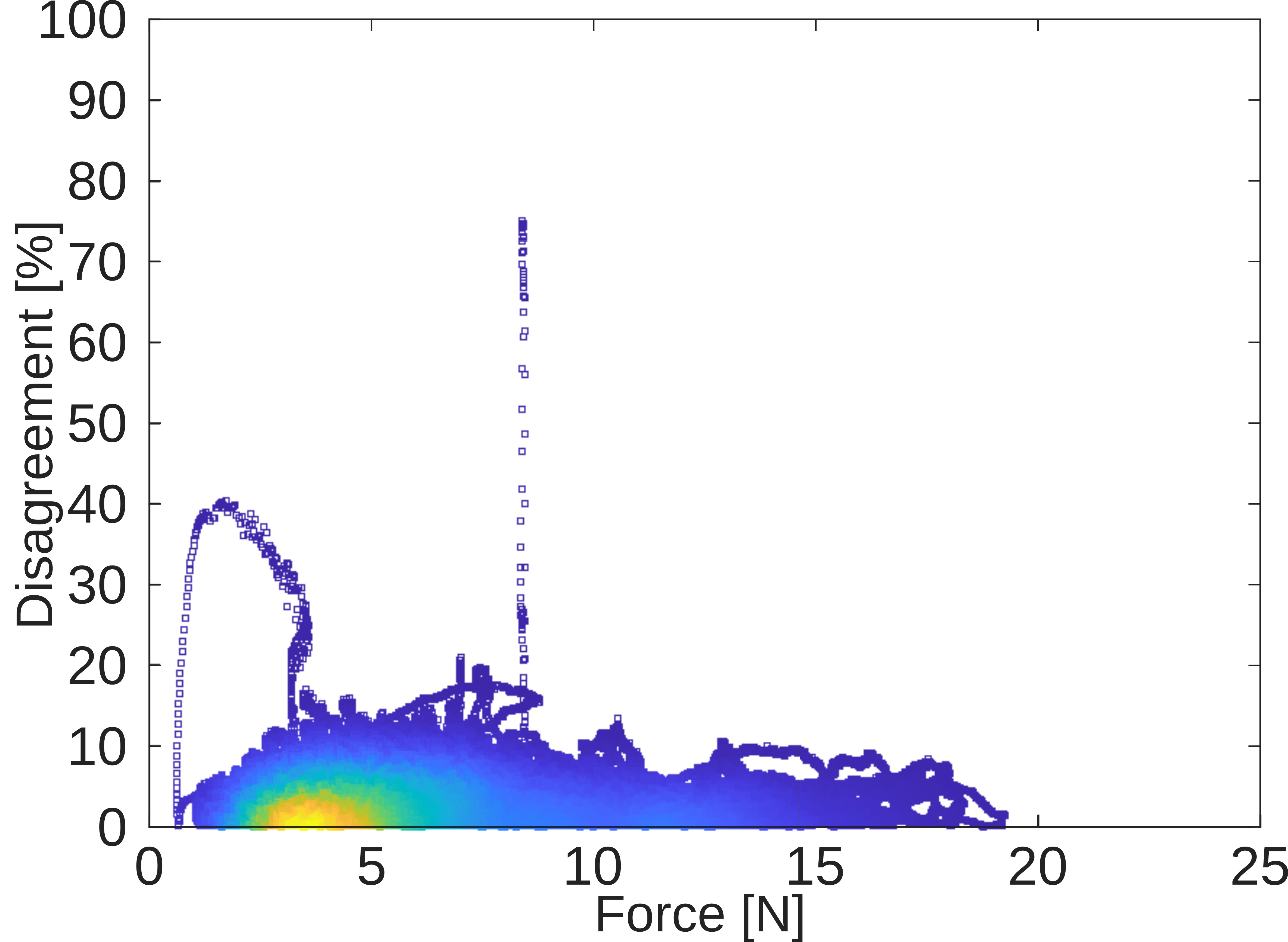}}
    \hfill
    \subfloat[\label{fig:scatterplots:shared:dis-eff}]{%
          \includegraphics[width=0.33\linewidth]{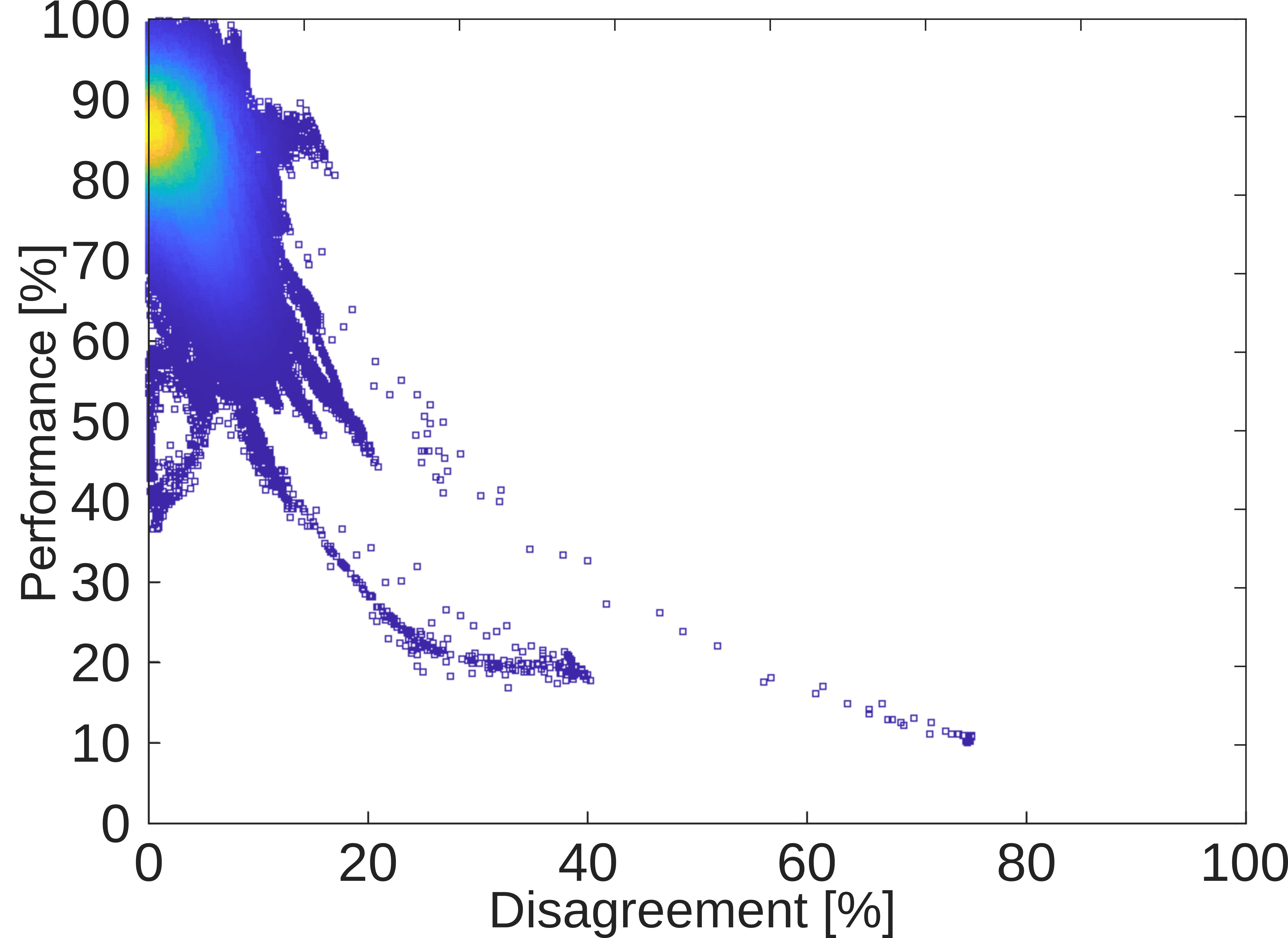}}
    \hfill \\
    \subfloat[\label{fig:scatterplots:imp:force-eff}]{%
          \includegraphics[width=0.33\linewidth]{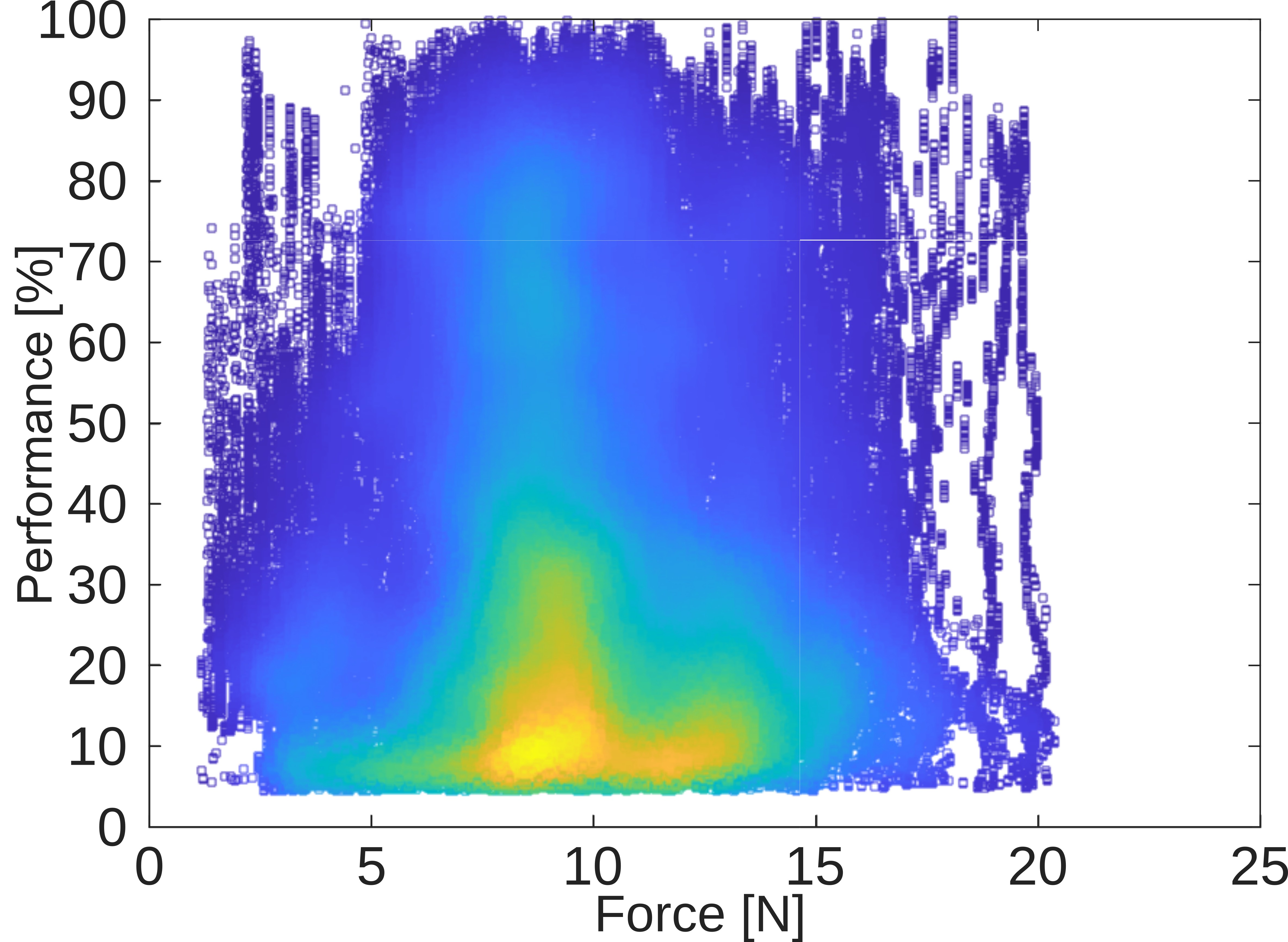}}
    \hfill
    \subfloat[\label{fig:scatterplots:imp:force-dis}]{%
          \includegraphics[width=0.33\linewidth]{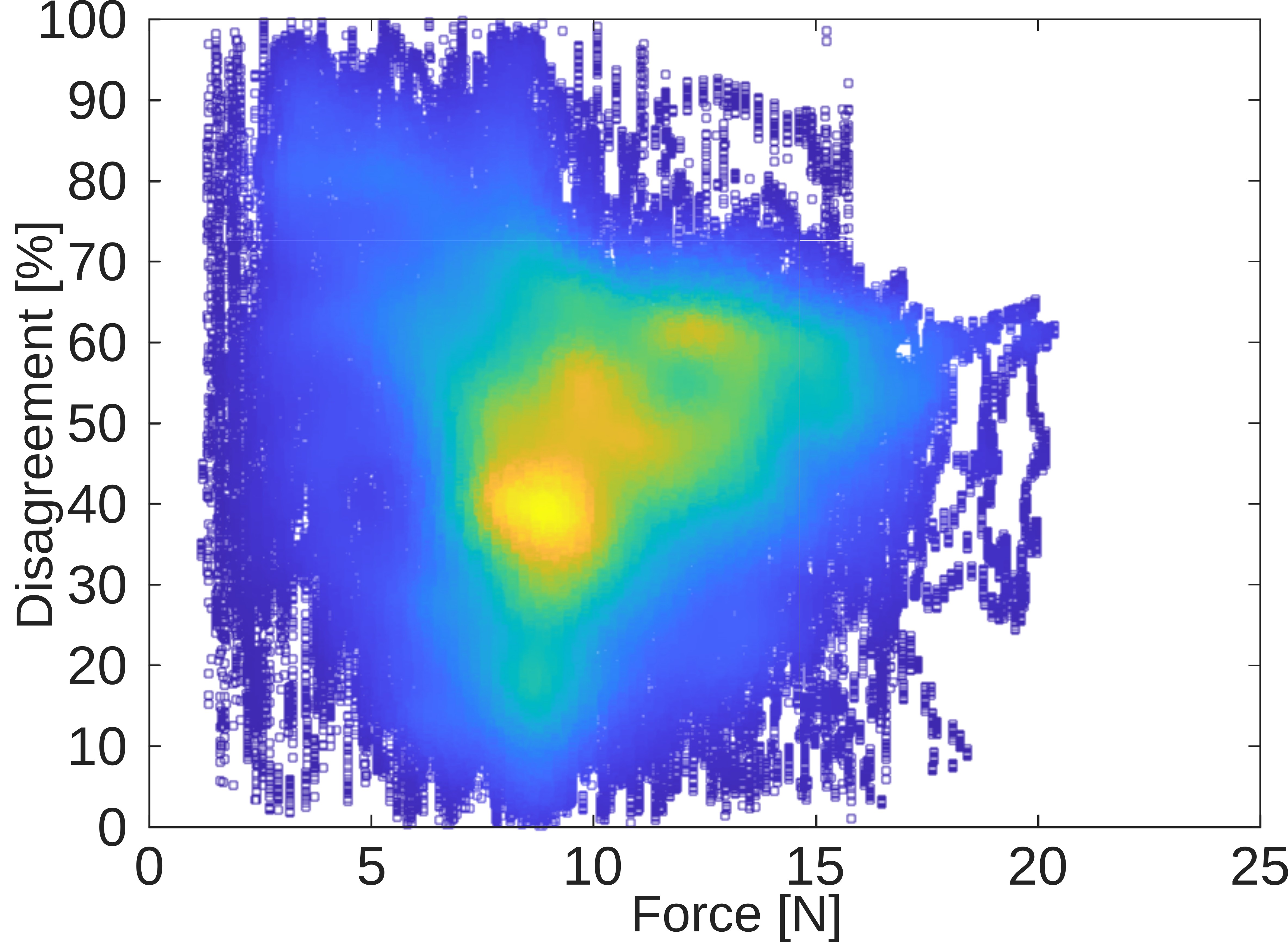}}
    \hfill
    \subfloat[\label{fig:scatterplots:imp:dis-eff}]{%
          \includegraphics[width=0.33\linewidth]{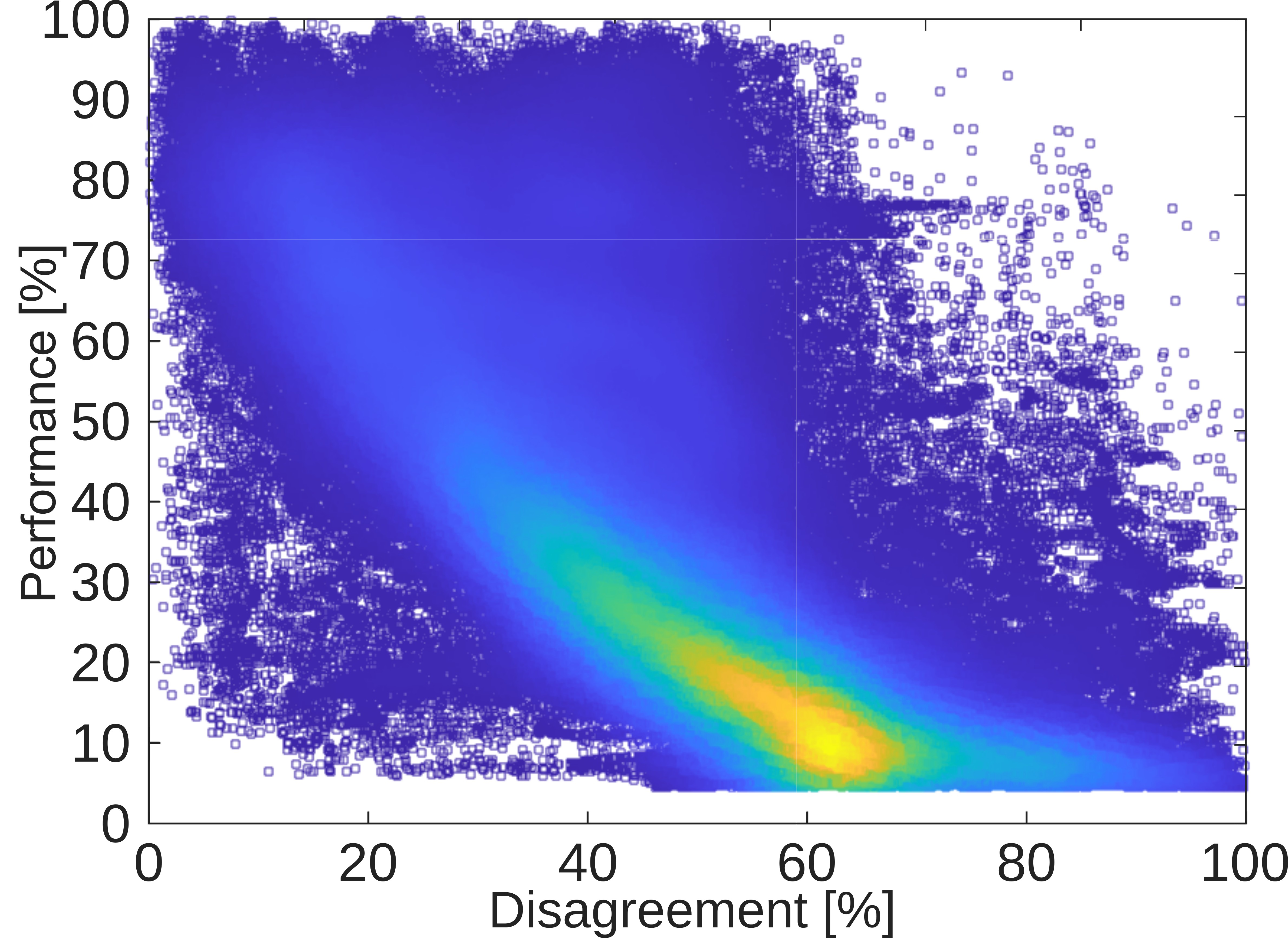}}
    \caption{Scatter plots of all Standalone (top row), SC (middle row), and IC (bottom row) tests comparing force vs. performance (left column); force vs. disagreement\protect\footnotemark (mid column); and disagreement vs. performance (right column). Light-colored areas are more likely to happen.}      
    \label{fig:scatterplots} 
\end{figure*}

Commented results show that some presented metrics could present dependencies. Hence, Figure \ref{fig:scatterplots} presents scatter plots between force, performance, and disagreement in StandAlone, SC and IC modes. To obtain these scatter plots, we have paired all values of the magnitude in the y-axis in each control mode with the corresponding values of the magnitude in the x-axis for all users at each point of their trajectories. A 2D-Histogram is applied to the dataset to find the most-likely pairs, which are displayed in brighter colors. These plots provide information on how correlated two magnitudes are (grouping) and which value pairs are most common. The standalone mode has no disagreement, so Figure~\ref{fig:scatterplots}.b can be used to study force values, which are a bit larger than in SC mode. 

Focusing on the dependency between performance and force, we can observe a tighter relationship in Standalone (Figure~\ref{fig:scatterplots}.a) and SC (Figure~\ref{fig:scatterplots}.d) when compared to IC mode (Figure~\ref{fig:scatterplots}.g). Standalone and SC modes always result in larger performance values for any given force, but this is not necessarily true in IC mode. 

Force and disagreement dependency is also affected by mode. Figure~\ref{fig:scatterplots}.h shows a loose relation between both (i.e., a large range of Disagreement values are possible for any given force), but the general trend is that a larger force results in a larger average force. 
SC mode has a stronger dependency: Figure~\ref{fig:scatterplots}.e presents a much narrower range for Disagreements, and most likely force values are mostly unaffected by Disagreement. 

The same effect can be appreciated in Figure \ref{fig:scatterplots}.c and Figure~\ref{fig:scatterplots}.f:
most likely, performance values in SC are a bit higher than in StandAlone. 
Also, we can see the inverse relationship between performance and disagreement in Figure~\ref{fig:scatterplots}.f and Figure~\ref{fig:scatterplots}.i.
And again, SC mode has a stronger dependency, with narrower - and higher- performance ranges for any given performance. 

Overall, it can be concluded that in SC, the trade-off between performance and disagreement is acceptable, and exerted forces are not too affected by assistance. Impedance control, however, affects relationships among performance, force, and disagreement significantly.

\subsection{Hypotheses Validation}

\footnotetext{Note that in Standalone control, the user command is equal to the shared command, so there is no disagreement. In this case, we have a 1-D Histogram instead of a 2-D scatterplot.}

To validate the different hypotheses in Section~\ref{sec:setup}, mainly if assistance improves performance on a need basis, ANOVA analysis has been used. Table~\ref{tab:anova} shows the p-values obtained for each hypothesis. $\Delta\eta_{hand}$ is used to designate the difference in performance between dominant and non dominant hand for each user, that is needed to validate H3.  
It can be noted that all hypotheses except H3 are validated by an ANOVA test. SC improves task performance (H1) and results in a better figure fit (H4). 
Also, average RMSPE and cognitive load (according to our proposed metrics) decrease in shared mode ($p$ under 5\% for H2 and H5). However, H3 is rejected, i.e., it can not be stated that SC homogenizes results from dominant to non-dominant hand for this population. Indeed, there is little difference in average performance between performance with the dominant and non-dominant hand when we assess the whole population. However, this result may mean that some users are equally good with both hands. As stated above, the proposed blend of SC is not meant to boost human performance to robot levels but to assist those users who need more help to achieve adequate performance. Hence, an additional p-value has been calculated for a restricted population to check hypothesis H3 only with people who present significant differences in hand performance ($\Delta \eta_{hand}$) ($H'_i$). 
H3' is checked over people with the largest differences in performance between dominant and non-dominant hand tests (upper tercile), i.e., people who use their dominant hand much better than the other one. 
In this case, H3' is confirmed: SC equalizes performance between left/right hands, significantly reducing performance differences. Hence, it could be concluded that SC provides help on a need basis.  

  \begin{table}[bthp]
  \caption{ANOVA Test results.}\label{tab:anova}
  \renewcommand{\tabcolsep}{1.5pt} 
    \centering
    \begin{tabular}{c c c c}
      \toprule
            Hyp.                  & Data                            & Factor            & ANOVA                  \\  \midrule 
            H1                    & user performance                & standalone/shared & 0.0143                 \\ 
            H2                    & average RMSPE                    & standalone/shared & 0.0423                 \\ 
            H3                    & $\Delta \eta_{hand}$ (per user) & standalone/shared & \cellcolor{Gray} 0.848 \\ 
            H3'                   & $\Delta \eta_{hand}$ (per user) & standalone/shared & 0.0421                 \\ 
            H4                    & RMSPE variance                   & standalone/shared & 0.0347                 \\ 
            H5                    & command variation               & standalone/shared & 0.0395                 \\ \bottomrule 
    \end{tabular}
  \end{table}

\section{Discussion}
\label{sec:results}

The proposed SC for grippers aims to assist humans moving their arms following a trajectory. It adapts to human input, so different people may present different results.
Our SC law continuously combines robot and human commands weighted by their local performance, estimated in terms of smoothness and directness.
Emergent commands tend to preserve continuity. Besides, resulting trajectories tend to be smooth and precise as a whole. Besides, some evaluation metrics not explicitly included in the control function, like RMSPE, human discrepancy with emerging commands, cognitive load and completion time.) return good values in our SC mode.

We have compared SC to human Standalone and to a classic IC in tests with several healthy volunteers. SC improves local performance and reduces RMSE variance and average, probably because it makes human commands steadier and increases resistance progressively the further users move from their intended trajectory. The most remarkable effect in SC is a significant decrease in performance variance among volunteers and also from dominant to non-dominant hand per volunteer, meaning that SC adapts to user’s needs. Performance homogenization was an intended effect of performance weighting, as the system is not expected to raise performance to robot level but to improve performance to good human standards.

Assistance in SC is well accepted, as Disagreement indicates that human and robot commands tend to be more aligned than in other modes, leading to good precision and task times. Although users were not informed of the control mode in tests they were operating, most volunteers reported feeling more comfortable in SC. Disagreement is largest in IC mode and many volunteers reported that control was jagged in this mode. Higher Disagreement was usually detected in specific areas, e.g., arm motion when the hand is closer to their bodies.

Increase in variance in Intervention Level along with a lower command variation also seems to indicate that SC reduces number of corrections. Corrections are often associated to higher cognitive effort, although more complex analysis would be required to extract conclusions. 

Finally, SC reduces variation in completion time. Visually, we observed that volunteers, especially those unfamiliar with robots, often applied too little force, probably for fear of damaging the robot. As the system is collaborative, lack of human force often leads to robot intermittent stops. Moreover, when the robot stopped, users often increased force sharply to overcome friction, just to quickly decrease it again when speed grew out of their comfort zone, leading to undesired force oscillations. This issue happened more often in StandAlone and IC modes but did not affect all users. A potential solution to this problem could be to assess forces required for specific tasks by different individuals to further adapt assistance to users and include this knowledge in control laws via learning user and/or task models.

\section{Conclusions}
\label{sec:Conclusion}
This paper has presented a new SC method for pHRI 
robots. The system is intended to allow continuous collaboration between humans and robots in physical contact for a common task. Emergent commands result from the combination of human and robot commands weighted by their respective local performance,  calculated in smoothness and directness. Different performance factors, like safety, could be used depending on the task. The reactive nature of the SC control law provides motion continuity, and performance weighting allows humans to contribute to tasks as much as possible, depending on their skills. This SC blend aims to assist on a need basis, i.e., if human performance is high enough, not much assistance is needed. As a result, emergent performance is homogenized despite standalone human skills. In presented tests, it was checked that SC also homogenized performance between dominant and non-dominant hands for volunteers.

Experiments conducted by volunteers show that SC performance increases, but mostly, that variance decreases significantly. Average tracking error and its variance also decrease in SC mode. Intervention Level shows that people continuously contribute to control, but low command variation (loosely) indicates that no extra cognitive load is involved. Volunteers reported different levels of comfort in tests, although most reported feeling most comfortable in a SC mode. In general, Disagreement changed depending on the relative position of the arm with respect to the volunteer's body, as some maneuvers are less comfortable than others, but, as a whole, it remains low in SC mode. It was also observed that volunteers with more skill showed higher average Disagreement in SC mode than other users. In all assessed metrics, SC outperformed Standalone and IC generally, although to different degrees. This conclusion was validated using ANOVA tests.

Future work will focus on including security constraints in the SC law to perform more complex collaborative maneuvers in the 3D space and on designing a better robot navigation strategy. We also contemplate adaptation of this SC approach to more critical scenarios where contact is initiated by the robot, such as rehabilitation, rescue, first aid, or nursing robots. In these scenarios, the robot may include additional goals, keeping the human safety concerns at all times but also adapting to human input, so performance factors would need to contemplate this, and further adaptation will be required.

\section*{Declaration of Competing Interest}
The authors declare that they have no known competing financial interests or personal relationships that could have appeared to influence the work reported in this paper.

\section*{Acknowledgements}
This work was supported by project PID2021-127221OB-100 from the Spanish Ministerio de Ciencia e Innovaci\'on and project UMA20-FEDERJA-052 from the University of M\'alaga.

\bibliographystyle{elsarticle-num}
\bibliography{biblio.bib}

\end{document}